%% file: paper_arXiv_2.tex
\documentclass[11pt]{article}
\usepackage{graphicx}
\usepackage{float}    
\usepackage{verbatim} 
\usepackage{amsmath, amssymb, amsthm,mathtools}  
\usepackage{subfig}   
\usepackage[colorlinks=true,citecolor=blue,linkcolor=black,urlcolor=magenta]{hyperref}
\usepackage{bookmark}
\usepackage{fullpage}
\usepackage{enumerate}
\usepackage{paralist}
\usepackage{xspace}
\usepackage{caption}
\usepackage{url}
\usepackage{mathrsfs}
\usepackage{algorithm}
\usepackage{algorithmic}
\usepackage{color}
\usepackage{microtype}
\usepackage[section]{placeins}
\usepackage{lscape}
\usepackage{multirow}
\input{math_commands.tex}

\usepackage{wrapfig}
\usepackage{dsfont}
\usepackage[utf8]{inputenc} 
\usepackage[T1]{fontenc}    
\usepackage{booktabs}       
\usepackage{multirow}
\usepackage{amsfonts}       
\usepackage{nicefrac}       
\usepackage{microtype}      
\usepackage{xcolor}         

\usepackage{amsmath, amssymb, amsthm,mathtools}
\usepackage{bm}
\usepackage{bbm}
\usepackage{natbib}

\usepackage{lipsum}
\usepackage{epstopdf}
\usepackage{ulem}

{\hspace*{\fill}$\Box$\par\vspace{4mm}}
{\hspace*{\fill}$\Box$\par}

\renewcommand{\emph}{\textit}

\newcommand{\cA}{\mathcal{A}}
\newcommand{\cB}{\mathcal{B}}

\newcommand{\cL}{\mathcal{L}}
\newcommand{\cM}{\mathcal{M}}

\newcommand{\norm}[1]{\left\lVert#1\right\rVert}

\newcommand{\bX}{\boldsymbol{X}}
\newcommand{\tbX}{\tilde{\boldsymbol{X}}}
\newcommand{\hbX}{\widehat{\boldsymbol{X}}}

\newcommand{\bY}{\boldsymbol{Y}}
\newcommand{\tbY}{\tilde{\boldsymbol{Y}}}

\newcommand{\bE}{\boldsymbol{E}}
\newcommand{\tbE}{\tilde{\boldsymbol{E}}}
\newcommand{\bD}{\boldsymbol{D}}

\newcommand{\bU}{\boldsymbol{U}}
\newcommand{\tbU}{\tilde{\boldsymbol{U}}}
\newcommand{\bV}{\boldsymbol{V}}
\newcommand{\bSigma}{\boldsymbol{\Sigma}}
\newcommand{\bDelta}{\boldsymbol{\Delta}}

\newcommand{\bA}{\boldsymbol{A}}
\newcommand{\tbA}{\tilde{\boldsymbol{A}}}

\newcommand{\bZ}{\boldsymbol{Z}}

\newcommand{\bw}{\boldsymbol{w}}
\newcommand{\bI}{\boldsymbol{I}}
\newcommand{\bL}{\boldsymbol{L}}

\newcommand{\bM}{\boldsymbol{M}}
\newcommand{\bG}{\boldsymbol{G}}
\newcommand{\bP}{\boldsymbol{P}}
\newcommand{\tbP}{\tilde{\boldsymbol{P}}}

\newcommand{\bx}{\boldsymbol{x}}
\newcommand{\by}{\boldsymbol{y}}
\newcommand{\bz}{\boldsymbol{z}}

\newcommand{\hkappa}{\hat{\kappa}}
\newcommand{\bv}{\boldsymbol{v}}

\newcommand{\supp}{\mathrm{supp}}

\newtheorem{theorem}{Theorem}

\newtheorem{lemma}{Lemma}

\newtheorem{assumption}{Assumption}

\title{Simple Alternating Minimization Provably Solves Complete Dictionary Learning
\thanks{Corresponding author: Salar Fattahi, \texttt{fattahi@umich.edu}\\
Financial support for this work was provided by NSF Award DMS-2152776, ONR Award N00014-22-1-2127, NSF CAREER Award ECCS-2047462, and
in part by the C3.ai Digital Transformation Institute.
}
}
\author{
	Geyu Liang\\
	Industrial and Operations Engineering\\
	University of Michigan\\
	\texttt{lianggy@umich.edu}
	\and
	Gavin Zhang\\
	Electrical and Computer Engineering\\
	University of Illinois at Urbana-Champaign\\
	\texttt{jialun2@illinois.edu}
	\and
	Salar Fattahi\\
	Industrial and Operations Engineering\\
	University of Michigan\\
	\texttt{fattahi@umich.edu}
	\and 
	Richard Y. Zhang\\
	Electrical and Computer Engineering\\
	University of Illinois at Urbana-Champaign\\
	\texttt{ryz@illinois.edu}
}

\begin{document}

\maketitle

\begin{abstract}
This paper focuses on the \textit{noiseless complete dictionary learning problem}, where the goal is to represent a set of given signals as linear combinations of a small number of atoms from a learned dictionary. There are two main challenges faced by theoretical and practical studies of dictionary learning: the lack of theoretical guarantees for practically-used heuristic algorithms and their poor scalability when dealing with huge-scale datasets. Towards addressing these issues, we propose a simple and efficient algorithm that provably recovers the ground truth when applied to the nonconvex and discrete formulation of the problem in the noiseless setting. We also extend our proposed method to mini-batch and online settings where the data is huge-scale or arrives continuously over time. At the core of our proposed method lies an efficient preconditioning technique that transforms the unknown dictionary to a near-orthonormal one, for which we prove a simple alternating minimization technique converges linearly to the ground truth under minimal conditions. Our numerical experiments on synthetic and real datasets showcase the superiority of our method compared with the existing techniques.
\end{abstract}

\section{Introduction}

The dictionary learning problem seeks to
represent data as a linear combination of a small number of basis
elements, known as \emph{atoms}, which are learned from the data itself.
The resulting collection of atoms, known as the \emph{dictionary},
can then be used for a variety of signal processing tasks \cite{mairal2007sparse,liu2018online}, often as
a plug-in replacement for classical bases based on cosines, wavelets,
or Gabor filters. The advantage of a learned dictionary over these classical bases is
that it is tuned to the input dataset, and can therefore provide a
sparser representation for each input signal that uses fewer
atoms.

The problem of learning a dictionary that provides the \emph{sparsest}
representation is non-convex and highly difficult to solve, both in
theory and in practice. In this paper, we focus on learning a \emph{complete}
dictionary, whose number of atoms matches the dimensionality of the
data, via the following optimization problem: 
\begin{equation}
\min_{\bX,\bD}\|\bY-\bD\bX\|_{F}^{2}+\zeta\|\bX\|_{0}.\tag{CDL}\label{objective}
\end{equation}
Here, $\bY\in\mathbb{R}^{n\times p}$ is a data matrix whose columns
are observed signals, and our goal is to find a \emph{square} dictionary
matrix $\bD\in\mathbb{R}^{n\times n}$ and corresponding sparse code
$\bX\in\mathbb{R}^{n\times p}$ to approximately represent $\bY$
while minimizing its $\ell_{0}$ pseudo-norm (denoted as $\|\cdot\|_{0}$)
which counts its number of non-zero elements.

\begin{figure*}
    \centering
    \includegraphics[width=0.9\textwidth]{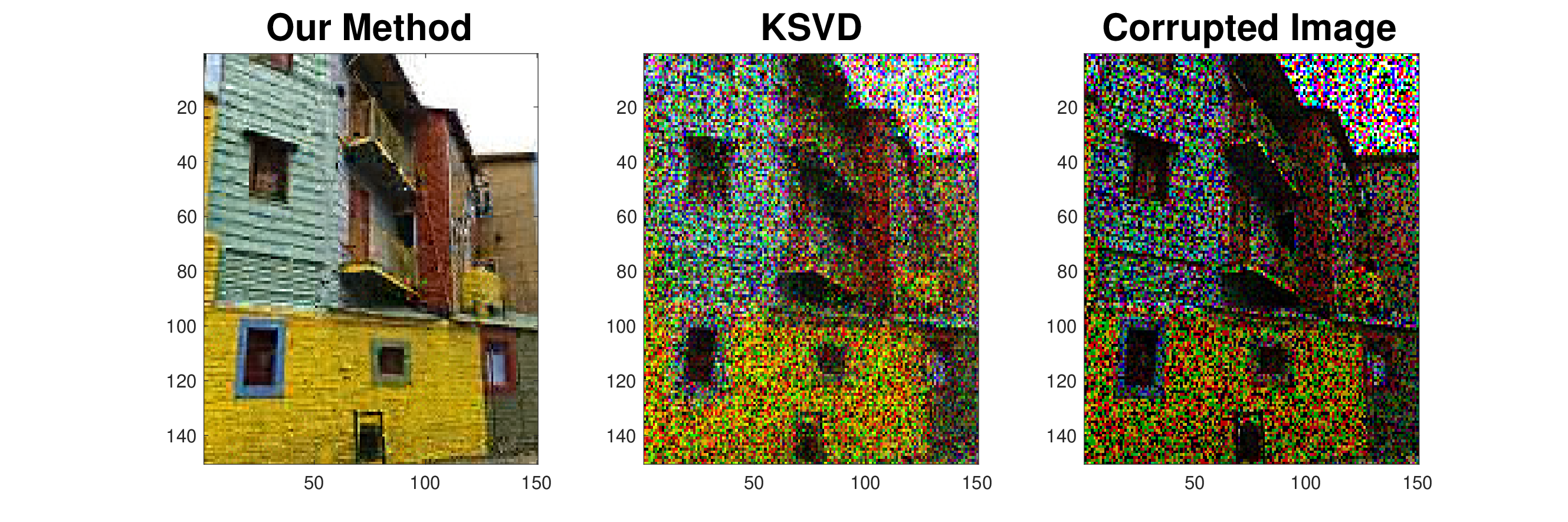}\vspace{-5mm}
    \vspace{-5pt} 
    \caption{\textbf{A comparison of image denoising using dictionaries learned via our proposed method (Algorithm \ref{alg: complete mini-batch}) and via KSVD}.  We choose a random landscape image and artificially corrupt 50\% of the pixels. Reconstruction is done via orthogonal matching pursuit with the learned dictionaries. 
    The corrupted original image is shown on the right, and the two reconstructed images are shown on the left. We see that the dictionary learned via our method achieves a much better denoising result than one learned via KSVD. We refer the readers to section 4.2 for the details of setup.}
    \label{fig:denoise}
\end{figure*}

The ability to solve \ref{objective} to optimality would allow us to strike the best-possible trade-off between the sparsity of the representation and its quality of fit (according to the preference expressed by the parameter $\zeta>0$).
Unfortunately, the $\ell_{0}$ pseudo-norm presents a significant difficulty, as it is not only non-convex, but also discrete and combinatorial. 
Even rigorously verifying that a learned dictionary is first-order locally optimal could require exhaustively enumerating all $2^{np}$ possible sparsity patterns. 
It is therefore understandable that the most widely-used heuristics like KSVD \cite{aharon2006k} and MOD \cite{engan1999method} do not actually guarantee convergence to an optimal solution, not even when provided with a near-optimal initial guess. 
Instead, the state-of-the-art for provable optimality relies on relaxing the combinatorial $\ell_{0}$ pseudo-norm into the convex $\ell_{1}$ norm \cite{agarwal2016learning} or the convex $\ell_{4}$ norm \cite{zhai2020complete}. 
Under assumptions like mutual incoherency~\cite{arora2014new}, restricted isometry property (RIP)~\cite{agarwal2016learning} or large sample complexity~\cite{sun2016complete1,zhai2020complete}, the relaxation is exact and an optimal solution to the original combinatorial problem can be recovered. However, when the relaxation is inexact, the recovered solution may be drastically different than the solution to the original \ref{objective}.

In this paper, we revisit the combinatorial $\ell_{0}$ pseudo-norm in \ref{objective}, motivated by the fact that
the \emph{orthogonal} instance (in which
the dictionary $\bD$ is constrained to be orthonomal) is significantly
easier to analyze. In the orthogonal setting, alternating
minimization is \emph{simple}, in that the minimizations over $\bD$
and $\bX$ at each iteration have simple closed-form solutions. We give the first proof that the resulting sequence converges to the ground truth model under minimal conditions, provided that the algorithm is initialized properly. To extend to general complete dictionaries, which are not necessarily orthogonal, our key idea is to use a data-driven preconditioning step to ``orthogonalize'' the data.
Surprisingly, this preconditioning step allows us to extend our linear convergence guarantee to general complete dictionary learning without significant modifications. As shown in Figure~\ref{fig:denoise}, the resulting preconditioned algorithm learns more powerful dictionaries that strike a better 
balance between the sparsity of representation and the quality of fit, 
than the commonly-used heuristic KSVD.

\subsection{Summary of results}\label{sec: summary of results}
At the heart of our method to recover a pair $(\bD^*,\bX^*)$ is an efficient alternating minimization
algorithm for the \emph{orthonormal} dictionary learning problem,
which reads 
\begin{equation}
\min_{\bX,\bD}\|\bY-\bD\bX\|_{F}^{2}+\zeta\|\bX\|_{0}\quad\text{ s.t. }\bD\in\mathbb{O}(n)\tag{ODL}\label{eq:odl}
\end{equation}
where $\mathbb{O}(n)$ denotes the set of $n\times n$ orthonormal
matrices. While the orthogonality constraint adds another nonconvex component to the original formulation, we point out that the minimization of $\bX$ for a
fixed orthonormal $\bD\in\mathbb{O}(n)$ has a cheap closed-form solution:
\begin{equation*}\label{eq: closed form 1}
    \operatorname{argmin}_{\bX}\|\bY-\bD\bX\|_{F}^{2}+\zeta\|\bX\|_{0}=\mathrm{HT}_{\zeta}(\bD^{T}\bY)
\end{equation*}
via a hard-thresholding operator $\mathrm{HT}_{\zeta}(\cdot)$ at
level $\zeta$, which is defined as: 
\[
\left(\mathrm{HT}_{\zeta}(\bA)\right)_{ij}=\begin{cases}
\bA_{ij} & \text{if}\quad|\bA_{ij}|\ge\zeta,\\
0 & \text{if}\quad|\bA_{ij}|<\zeta.
\end{cases}
\]
Similarly, the minimization of an orthonormal $\bD\in\mathbb{O}(n)$
for a fixed $\bX$ also has a well-known closed-form solution via
the so-called orthogonal Procrustes problem 
\begin{equation*}\label{eq: closed form 2}
    \underset{\bD\in\mathbb{O}(n)}{\operatorname{argmin}}\ \|\bY-\bD\bX\|_{F}^{2}+\zeta\|\bX\|_{0}=\mathrm{Polar}(\bY\bX^{\top}),
\end{equation*}
where $\mathrm{Polar}(\bA)=\bU_{\bA}\bV_{\bA}^{\top}$, and $\bU_{\bA}\bSigma_{\bA}\bV_{\bA}^{\top}$
is the Singular Value Decomposition (SVD) of $\bA$. Therefore, the
following iterations 
\begin{equation}
\bX^{(t+1)}=\mathrm{HT}_{\zeta}(\bD^{(t)\top}\bY),\qquad\bD^{(t+1)}=\mathrm{Polar}\left(\bY\bX^{(t+1)\top}\right)\label{eq:altmin}
\end{equation}
correspond exactly to an alternating minimization solution of Problem
\ref{eq:odl}. 
While (\ref{eq:altmin}) has been derived before, we
give the first rigorous proof that the sequence locally converges
at a linear rate with minimal conditions on the sparsity rate of $\bX^{*}$. Concretely, we prove that when the data are generated
as $\bY=\bD^{*}\bX^{*}$, with $\bD^{*}$ an orthogonal dictionary,
and $\bX^{*}$ a sparse random matrix, (\ref{eq:altmin}) converges
linearly to a ground truth $(\bD^{*},\bX^{*})$, when it is initialized
within an $O\left(\frac{1}{\sigma\sqrt{\theta n}}\right)$ Frobenius neighborhood of this solution.

We use a data-driven preconditioning step to generalize (\ref{eq:altmin})
to general instances of \ref{objective}, for which the square
dictionary matrix $\bD^*$ is not necessarily orthonormal. The basic
idea is to learn an orthonormal dictionary $\tilde{\bD}$ with respect
to the preconditioned data matrix $\tilde{\bY}=(\bY\bY^{\top})^{-1/2}\bY$,
and then to output a complete dictionary $\bD=(\bY\bY^{\top})^{+1/2}\tilde{\bD}$
by reversing the preconditioning. Indeed, if the data are generated
as $\bY=\bD^{*}\bX^{*}$ with $\bD^{*}$ deterministic, and $\bX^{*}$
a sparse random matrix satisfying $\mathbb{E}[\bX^{*}\bX^{*\top}]=I_{n}$,
then $\mathbb{E}[\bY\bY^{\top}]=\bD^{*}\mathbb{E}[\bX^{*}\bX^{*\top}]\bD^{*\top}=\bD^{*}\bD^{*\top}$.
Therefore, the preconditioned data are generated as $\tilde{\bY}\approx\tilde{\bD}^{*}\bX^{*}$
with respect to an orthogonal dictionary $\tilde{\bD}^{*}=(\bD^{*}\bD^{*\top})^{+1/2}\bD^{*}$
and the same sparse code $\bX^{*}$. We show that such preconditioning can be efficiently implemented via low-rank updates of the corresponding Cholesky factors. One of our main contributions is to show that this simple preconditioning step 
allows us to extend our linear convergence guarantee from orthogonal dictionaries (\ref{eq:odl}) to general complete dictionaries (\ref{objective}), even in the online and mini-batch settings.

A key strength of our convergence guarantee is that it does not rely on any incoherency/ RIP
assumptions. In fact, we can recover \textit{very ill-conditioned} dictionaries
that are highly coherent, although at the expense of a higher sample
complexity. Moreover, we can recover codes with sparsity levels in
the order of $O(n)$; in contrast, existing methods based
on alternating minimization can handle sparsity levels of at most
$O(n^{\alpha})$ for some $\alpha\leq1/2$. 
Another technique that achieves a near-linear sparsity level $O(n^{1-\gamma})$
for $\gamma>0$ is based on \textit{sum-of-squares hierarchy}~\cite{barak2015dictionary},
which is indeed inefficient in practice. Using Riemannian optimization techniques, \cite{sun2015complete} was the first to show that \ref{objective} with linear sparsity level can be solved in polynomial time. However, the Riemannian optimization algorithm can be expensive, and the associated sample complexity has a dependency of $\Omega(n^5)$ for orthogonal dictionaries and $\Omega(n^7)$ for complete dictionaries \cite{sun2015complete}.

In practice, the iterations (\ref{eq:altmin}) admit highly efficient
implementations, that can fully take advantage of the massive parallelism
inherent in modern hardware. The hard-threshold operator is embarrassingly
parallel, while the SVD operation that constitutes the polar operator
can be implemented using hardware-optimized LAPACK implementations.
The most expensive part of (\ref{eq:altmin}) is actually its need
to iterate and sum over all $p$ columns of the data matrix $\bY$
at every iteration. To address potential scalability issues with a
very large $p$, we propose a mini-batch version, 
\begin{equation}
\bX_{(\cdot,i)}^{(t+1)}=\mathrm{HT}_{\zeta}(\bD^{(t)\top}\bY_{(\cdot,i)})\quad\forall i\in\Omega,\qquad\bD^{(t+1)}=\mathrm{Polar}\left(\sum_{i\in\Omega}\bY_{(\cdot,i)}\bX_{(\cdot,i)}^{(t+1)\top}\right).\label{eq:saltmin}
\end{equation}
that approximates $\bY\bX^{(t+1)\top}\approx\sum_{i\in\Omega}\bY_{(\cdot,i)}\bX_{(\cdot,i)}^{(t+1)\top}$
over a mini-batch $\Omega\subseteq\{1,2,\dots,p\}$, and updates only
the sparse codes $\bX_{(\cdot,i)}^{(t+1)}$ associated with the mini-batch
samples $i\in\Omega$. We show that
the preconditioner can also be incrementally updated,
via an efficient low-rank update formula for the Cholesky factor,
which is comparable to the classical Sherman-Morrison-Woodbury
formula for updating the determinant/matrix inverse. We show that the accuracy of the recovered dictionary is explicitly lower bounded by the statistical error of the preconditioner. We also show that such an error will diminish in the online setting as more samples come in. This is the first time that efficient updating is applied to methods that use the preconditioning for \ref{objective}.

Despite its mini-batch nature, we show that
our method enjoys the same sharp linear convergence to the true solution.
Finally, we extend our technique to the \textit{online dictionary
learning}, where the data samples arrive sequentially over time. To
the best of our knowledge, existing methods
for online dictionary learning work with batch sizes of at least $O(n)$
(see e.g.~\cite[Theorem 1]{rambhatla2019noodl} and~\cite[Theorem 2]{arora2015simple}).
In contrast, our proposed preconditioner admits an efficient triangular
rank-one update with as few as one new sample, thereby making it particularly
appealing in the online setting.

\subsection{Related work}\label{sec: related work}

\paragraph{Optimality conditions} As the first question, we must first verify whether a  ground truth $(\bD^*,\bX^*)$ is a global minimizer of \ref{objective} for the generative model $\bY=\bD^*\bX^*$. Of course, if this were not the case, solving \ref{objective}---even to global optimality---may not be enough to recover a ground truth $(\bD^*,\bX^*)$. A series of work  \cite{gribonval2010dictionary,geng2014local,jenatton2012local,gribonval2015sparse,schnass2015local,cohen2019identifiability,hu2023global} studied the local optimality of the ground truth for \ref{objective} by replacing $\ell_0$ norm with $\ell_1$ norm. The work \cite{spielman2012exact} elegantly showed that the ground truth is the unique global minimizer to \ref{objective} when $\zeta\rightarrow 0$. Specifically, when $p=\Omega(n\log n)$, they showed that any ground truth $(\bD^*,\bX^*)$ is the global minimizer to $\min_{\bX,\bD} \|\bX\|_0$ subject to the constraint $\bY=\bD\bX$. The question remains to be answered: how to design a provable algorithm for recovering $(\bX^*, \bD^*)\in \cM$.

\paragraph{Alternating minimization} The empirical success achieved by methods like MOD and KSVD has encouraged the emergence of many alternating minimization algorithms \cite{lee2006efficient,mairal2009online,bao2014l0}. Towards providing a provable guarantee for alternating minimization, it is common to replace the $\ell_0$ pseudo-norm with its convex surrogate $\ell_1$ norm \cite{agarwal2016learning, chatterji2017alternating, malezieux2021understanding, tolooshams2021stable}. Such a compromise is due to the prohibitive computational cost and formidable analytical challenges of $\ell_0$ pseudo-norm. However, $\ell_1$ relaxation is biased towards solutions with smaller entries, thereby leading to an inferior sparsity level \cite{zhang2010nearly}. On the algorithmic side, methods based on $\ell_1$ relaxation need to solve variants of the LASSO problem at each iteration either exactly \cite{agarwal2016learning, chatterji2017alternating} or approximately via an automatic differentiation with backpropagation \cite{malezieux2021understanding, tolooshams2021stable}. In this work, we focus on the formulation of \ref{objective} without any convex relaxation, which is the original intention of sparse dictionary learning. Unlike the $\ell_1$ method, the theoretical underpinnings of the alternating algorithms for the $\ell_0$ formulation are far less explored. In 
\cite{schnass2018convergence, hu2023global,schnass2020compressed, pali2023dictionary}, the authors study the convergence behavior of a variant of K-SVD. Specifically, the dictionary update step is performed by maximizing the absolute norm of the $S$-largest responses where $S$ is the number of non-zero entries in each signal. In \cite{ravishankar2020analysis}, authors propose an alternating scheme based on sorting the non-zero entries of sparse codes. However, the success of these algorithms relies on restrictive generative models like symmetric decaying \cite{schnass2015local} and fixed sparsity $S$ \cite{ravishankar2020analysis}.

\paragraph{Riemannian manifold optimization for complete dictionary learning} 
One line of work focuses on solving \ref{objective} via Riemannian manifold optimization techniques \cite{qu2014finding,zhai2020complete}. Notably, the work~\cite{sun2016complete1} was the first to show that a smoothed variant of \ref{objective} based on $\ell_1$ relaxation is devoid of spurious local solutions. Similar benign landscape results have also been independently discovered in the analysis of other problems like matrix factorization~\cite{ge2017no}. As a result, one can provably recover one dictionary atom at a time via the Riemannian trust-region method. The work \cite{gilboa2019efficient} showed that similar results can be achieved by first-order methods on the Riemannian manifold. In a different approach, the work \cite{zhai2020complete} studied the $\ell_4$ maximization on the Steifel manifold, which is further generalized in \cite{shen2020complete} to $\ell_p$ maximization. Akin to our own work, preconditioning the data matrix plays an important role in these algorithms.

\paragraph{Online dictionary learning} Another line of research has focused on the online variants of \ref{objective}. The work \cite{mairal2009online} first proposed an online algorithm that solves $\ell_1$ relaxation of \ref{objective}, but it only guarantees convergence to a critical point. The work \cite{lu2013online} enhanced the usability and practicality of 
$\ell_1$ relaxation but provided no theoretical guarantees. The work \cite{arora2015simple} proposed an alternating algorithm that can recover the ground truth $(\bX^*,\bD^*)$, which is further improved in \cite{rambhatla2019noodl}. However, both variants have a specific requirement on the size of the input for each iteration and cannot deal with situations where samples are received one by one.

\subsection{Notation}
We use $\bA_{(i,\cdot)}$, $\bA_{(\cdot,j)}$ and $\bA_{ij}$ to denote the $i$th row, $j$th column and $(i,j)$-th entry of $\bA$, respectively. We use $\|\bA\|_2$ to denote the spectral norm of $\bA$, $\|\bA\|_F$ to denote Frobenius norm of $\bA$, $\|\bA\|_{1,2}$ to denote the maximum $2$-norm of the columns of $\bA$, $\|\bA\|_0$ to denote the total number of non-zero entries in $\bA$, and $\|\bA\|_1$ to denote the entry-wise $\ell_1$ norm of $\bA$. The symbol $\bI_d$ denotes the $d\times d$ identity matrix, and $\mathbb{O}(n)$ denotes the orthogonal group in dimension $n$. We use $\supp(\bA)$ to refer to the set of indices of non-zero entries of $\bA$. 
The symbol $\sigma_i(\bA)$ denotes the $i$th largest singular value of $\bA$ and $\kappa(\bA)$ denotes the condition number of $\bA$. Given a matrix $\bA$, its polar decomposition is defined as $\mathrm{Polar}(\bA)=\bU_{\bA} \bV_{\bA}^\top$, where $\bU_{\bA} \bSigma_{\bA} \bV_{\bA}^\top$ is the SVD of $\bA$. 
We also define the operator $\mathcal{L}(\bA)=\bL_{\bA}^\top$, where $\bL_{\bA}\bL_{\bA}^\top$ is the Cholesky factorization of a positive semidefinite matrix $\bA$. 
For an event $\mathcal{E}$, its indicator function is denoted as $\mathbbm{1}_{\mathcal{E}}$. 
Given two sequences $f(n)$ and $g(n)$, the notations $f(n)\lesssim g(n)$ and $f(n)=O(g(n))$ imply that there exists a universal constant $C>0$ satisfying $f(n) \leq Cg(n)$ for all large enough $n$. Similarly, the notations $f(n)\gtrsim g(n)$ and $f(n)=\Omega(g(n))$ imply that there exists a universal constant $C>0$ satisfying $f(n) \geq Cg(n)$ for all large enough $n$. We use $f(n)=\omega(g(n))$ if for all constants $C>0$ we have $f(n) \geq Cg(n)$ for all large enough $n$. We say an event happens \textit{with high probability} if it occurs with probability of at least $1-n^{-\omega(1)}$ with respect to all randomness in the problem. 

\section{Our Method}\label{sec: our method}
We consider a noiseless model, where the data matrix is generated according to $\bY=\bD^*\bX^*$. For any signed permutation matrix $\Pi$, the pair $(\bD^*\Pi,\Pi^\top\bX^*)$ is also a valid ground truth that satisfies $\norm{\Pi^\top\bX^*}_0 = \norm{\bX^*}_0$.\footnote{A signed permutation matrix is a generalized permutation matrix whose nonzero entries are $\pm 1$.} Let $\cM = \{(\bD^*\Pi,\Pi^\top\bX^*): \text{$\Pi$ is a signed permutation}\}$. Our goal is to recover an arbitrary element $(\bD^*\Pi,\Pi^\top\bX^*)\in \cM$.

\subsection{Orthogonal Dictionary Learning}
We first assume that the dictionary is an orthogonal matrix $\bD^*\in\mathbb{O}(n)$. 
This leads to the following alternating minimization algorithm based on the closed-form solutions \eqref{eq:altmin}:
\begin{algorithm}[H]
	\caption{Alternating minimization for~\ref{eq:odl}}
	\label{alg: ortho offline}
	\begin{algorithmic}[1]
		\STATE{{\bf Input:} $\bY$, $\bD^{(0)}$, $\zeta$}
		\FOR{$t = 0,1,\dots, T$}
		\STATE{Set $\bX^{(t)} = \mathrm{HT}_{\zeta}\left(\bD^{(t)\top}\bY\right)$}
		\STATE{Set $\bD^{(t+1)} = \mathrm{Polar}\left(\bY \bX^{(t)\top}\right)$}
		\ENDFOR
		\RETURN{$\bD^{(T)}$,$\bX^{(T)}$}
	\end{algorithmic}
\end{algorithm}
We note that Algorithm~\ref{alg: ortho offline} has been studied before. The paper \cite{bao2013fast} reports the empirical performance of Algorithm~\ref{alg: ortho offline} on an image restoration task. The paper \cite{ravishankar2020analysis} provides a theoretical analysis for a variant of Algorithm~\ref{alg: ortho offline} based on sorted thresholding. One may even argue that Algorithm~\ref{alg: ortho offline} is similar to the Method of Optimal Directions (MOD). However, the existing theoretical guarantees for Algorithm~\ref{alg: ortho offline}  are indeed restrictive (see the discussion on alternating minimization in Section~\ref{sec: related work}). To bridge this knowledge gap, we first introduce our assumption on the sparse matrix $\bX$.
\begin{assumption}[Model for sparse code]\label{assump:sparse and normalization}
	The sparsity pattern of ground truth $\bX^*$ follows a Bernoulli distribution with parameter $\theta=(0,1]$. In particular,
	\begin{equation}\nonumber
	    \mathbbm{1}_{\bX^*_{ij}\ne 0}=B_{ij},\quad\text{where}\quad B_{ij}\overset{i.i.d.}{\sim}\mathcal{B}(\theta) \text{ for all $1\le i\le n$, $1\le j \le p$}.
	\end{equation}
    The non-zero values of $\bX^*$ are independently drawn from a sub-Gaussian distribution with mean zero and constant variance $\sigma^2$. Moreover, the magnitudes of non-zero entries of $\bX^*$ are lower bounded by some constant $\Gamma$. More specifically, we have
    \begin{equation}\nonumber
        |\bX^*_{ij}|\ge \Gamma,
        \quad\mathbb{E}(\bX^*_{ij})=0,\quad\text{and}
        \quad\mathbb{E}(\bX^{*2}_{ij})=\sigma^2, \text{  for every $(i,j)\in \supp(\bX^*)$.}
    \end{equation}
\end{assumption}
Now, we are ready to show the convergence of Algorithm~\ref{alg: ortho offline}:
\begin{theorem}\label{theorem: ortho offline}
    Suppose that $\bY = \bD^* \bX^*$ where $\bD^*$ is orthogonal and $\bX^*$ satisfies Assumption \ref{assump:sparse and normalization} with sparsity level $0 < \theta < 1/2$. Suppose that the initial dictionary $\bD^{(0)}$ satisfies $\|\bD^{(0)}-\bD^*\|_F \lesssim \frac{1}{\sigma\sqrt{\theta n}} $ for some $(\bD^*, \bX^*)\in \cM$. Moreover, suppose that $ n/\theta^2\lesssim p\lesssim n^\gamma$ for some constant $\gamma>0$. Then, with probability at least $1-n^{-\omega(1)}$, for any $T\ge 1$, the output of Algorithm~\ref{alg: ortho offline} with $\zeta = \Gamma/2$ satisfies:
    \begin{align}
        \|\bD^{(T)}-\bD^*\|_F \le  0.9^T\|\bD^{(0)}-\bD^*\|_F, \quad
        \|\bX^{(T)}-\bX^*\|_F \le  0.9^T\|\bX^{(0)}-\bX^*\|_F\nonumber
    \end{align}
\end{theorem}
Theorem~\ref{theorem: ortho offline} improves upon the existing results on two fronts: 
\paragraph{Linear sparsity level} We allow a constant fraction of entries in $\bX^*$ to be non-zero, thereby improving upon the best known sparsity level of $O(\sqrt{n}/\log n)$ for alternating minimization~\cite{arora2015simple}. Moreover, the imposed upper bound on the sample size asserts that it must be bounded by a polynomial function of $n$. This mild assumption is only included to simplify the presentation of our main result.

\paragraph{Linear sample complexity} In order to recover $\bD^*$ exactly, we only need to observe $O(n)$ many samples, which is $\log(n)$ factor smaller than the sample complexity required for the uniqueness of the solution when $\xi\rightarrow 0$ \cite{spielman2012exact}. This sample complexity is optimal (modulo constant factors) since it is impossible to recover the true dictionary with a sublinear number of samples even if $\bX^*$ is known. Note that the theoretical convergence of Algorithm~\ref{alg: ortho offline} is contingent upon a good initial dictionary. As shown in Lemma~\ref{lemma: exact support recovery}, the imposed condition on the initial dictionary automatically guarantees the recovery of the support for any $\bX^{(t)}$ with $t\ge 0$. In Subsection~\ref{subsec_init}, we discuss possible ways to obtain such an initial dictionary in theory and practice.
We also suspect that our initialization requirement can be improved to $\|\bD^{(0)}-\bD^*\|_{1,2}\le O(1/\log n)$ with fresh samples at every iteration, which is the best-known radius for alternating minimization \cite{arora2015simple}.

Despite its desirable properties, Algorithm~\ref{alg: ortho offline} suffers from two fundamental limitations. First, its convergence depends on the orthogonality of the true dictionary, which may not be satisfied in many applications. Second, it does not readily extend to huge-scale or online settings, where it is prohibitive or even impossible to process all data samples together. To address these challenges, we next extend our algorithm to complete (non-orthogonal) dictionary learning with mini-batch and online data.

\subsection{Mini-batch Complete Dictionary Learning}
To distinguish from \ref{eq:odl}, we denote the ground truth dictionary as $\bA^*$ (i.e., $\bY=\bA^*\bX^*$). Towards dealing with large $p$, we sub-sample from the columns of $\bY$. To address the non-orthogonality of the dictionary, we consider a preconditioner $\bP$ defined as 
\begin{equation}\nonumber
    \bP = \mathcal{L}\left(\left(\frac{1}{p\theta \sigma^2}\bY \bY^\top\right)^{-1}\right).
\end{equation}
Using this preconditioner, we obtain a new (preconditioned) data matrix $\tbY=\bP\bY$. To explain the intuition behind this choice of preconditioner, note that  $\frac{1}{p\theta \sigma^2}\bY \bY^\top\approx\bA^*\bA^{*\top}$ for large enough $p$, which allows us to write $\tbY\approx \mathcal{L}\left(\left(\bA^*\bA^{*\top}\right)^{-1}\right)\bA^*\bX^*$. Upon defining $\bD^* = \mathcal{L}\left(\left(\bA^*\bA^{*\top}\right)^{-1}\right)\bA^*$, one can check that $\bD^*\in \mathbb{O}(n)$ and $\tbY\approx\bD^*\bX^*$. Indeed, this is an instance of ODL and can be solved via Algorithm~\ref{alg: ortho offline}. The detailed implementation is provided in Algorithm~\ref{alg: complete offline}. 
We note that the dependency of the preconditioner on $\theta$ and $\sigma^2$ is purely to simplify the presentation and can be replaced by a simple normalization step in practice.
\begin{algorithm}
	\caption{Alternating minimization for mini-batch CDL}
	\label{alg: complete offline}
	\begin{algorithmic}[1]
		\STATE{{\bf Input:} $\bY$, $\bA^{(0)}$, $\zeta$}
		\STATE{Set $\bP\! =\! \mathcal{L}\left(\left(\frac{1}{p\theta \sigma^2}\bY \bY^\top\right)^{-1}\right)^\top$,\! $\tbY\! =\! \bP\bY$, and $\bD^{(0)} \!=\!  \bP\bA^{(0)}$.}
		\FOR{$t = 0,1,\dots, T-1$}
		\STATE{Sample $\tilde{p}$ many columns from $\tbY$ to be $\tbY^{(t)}$.}
		\STATE{Set $\tbX^{(t)} = \mathrm{HT}_{\zeta}\left(\bD^{(t)\top}\tbY^{(t)}\right)$.}
		\STATE{Set $\bD^{(t+1)} =  \mathrm{Polar}\left(\tbY^{(t)} \tbX^{(t)\top}\right)$.}
		\ENDFOR
		\RETURN{$\bA^{(T)} = \bP^{-1}\bD^{(T)}$ as an approximation to $\bA^*$}
	\end{algorithmic}
\end{algorithm}
Our next theorem characterizes the performance of Algorithm~\ref{alg: complete offline}. 

Without loss of generality, we assume that $\|\bA^*\|_2=1$ and set $\hkappa$ to be the condition number of $\bA^*$.
\begin{theorem}\label{theorem: complete offline}
    Suppose that $\bY = \bA^* \bX^*$, where $\bX^*$ satisfies Assumption \ref{assump:sparse and normalization} with sparsity level $0<\theta < 1/2$. Suppose that the initial dictionary $\bA^{(0)}$ satisfies $\|\bA^{(0)}-\bA^*\|_F\lesssim \frac{1}{\hkappa\sigma\sqrt{\theta n}}$ for some $(\bA^*, \bX^*)\in \cM$. Moreover, suppose that $n/\theta^2 \lesssim \tilde{p} \lesssim n^\gamma$, $\hkappa^{12}n^{3}\log^2 \tilde{p}/\theta\lesssim p$, $T \lesssim n^\beta$ and $\log n\gtrsim\beta+\gamma$, for some constants $\gamma, \beta>0$. Then with probability at least $1-n^{-\omega(1)}$, the output of Algorithm~\ref{alg: complete offline} with  $\zeta=\Gamma/2$ satisfies:
    \begin{align}
        \left\|\bA^{(T)}-\bA^*\right\|_F\le &\ 0.9^T\|\bA^{(0)}-\bA^*\|_F +O\left(\frac{n\hkappa^6}{\theta\sqrt{p}}\right).
    \end{align}
\end{theorem}
Theorem~\ref{theorem: complete offline} states that Algorithm~\ref{alg: complete offline} converges linearly to the true dictionary up to a statistical error $O\left(\frac{n\hkappa^6}{\theta\sqrt{p}}\right)$. This statistical error is due to the deviation of the preconditioner from its expectation, which diminishes with $p$.  A key distinction of our result is that we do not impose any incoherency requirement or restricted isometry property on $\bA^*$, which are common assumptions in existing results. We also highlight that the upper bound on $T$ is indeed very mild. To demonstrate this, note that, in order to guarantee $0.9^T\norm{\mathbf{A}^{(0)}-\mathbf{A}^*}_F = O\left(\frac{n\hkappa^6}{\theta\sqrt{p}}\right)$, it suffices to have $T=\Omega\left(\frac{\theta\sqrt{p}}{\hat\kappa^6n\log n}\right)$, a condition that satisfies $T=O(n^\beta)$ for some $\beta>0$ for all practical purposes.

\subsection{Online Dictionary Learning}
Even though the computational efficiency of Algorithm~\ref{alg: complete offline} is largely improved compared to its full-batch counterpart, it may not be implementable in huge-scale or online settings due to the dependency of the preconditioner on the entire dataset with size $p$. 
Both situations call for a cheap method to update the preconditioner more efficiently in an online fashion.

To update $\bP$ with a new sample $\by$, a naive approach would be to re-compute it from scratch which would cost $O(n^3)$ operations. However, we show that $\bP$ can be updated more efficiently in $O(n^2)$ operations by taking advantage of the more efficient rank-one updates on matrix inversion and Cholesky factorization. To this goal, we first use the Sherman-Morrison formula to update $\left(\bY\bY^\top\right)^{-1}$ as
\begin{align}
\left(\bY\bY^\top\!+\!\by\by^\top\right)^{-1} = &\left(\bY\bY^\top\right)^{-1}\!-\!\bv\bv^\top,\ \ \text{where}\ \ \bv = \left(1+\by^\top \left(\bY\bY^\top\right)^{-1}\by\right)^{-1/2}\left(\bY\bY^\top\right)^{-1}\by,\nonumber
\end{align}

which, given $\left(\bY\bY^\top\right)^{-1}$, can be obtained in $O(n^2)$ operations.
Given the above rank-one update for the inverse, the Cholesky factor $\cL\left(\left(\bY\bY^\top+\by\by^\top\right)^{-1}\right)$ can be obtained within $O(n^2)$ operations by performing triangular rank-one updates on $\cL\left(\left(\bY\bY^\top\right)^{-1}\right)$~\cite{krause2015more}. We explain the implementation of this method in Appendix~\ref{app_cholesky}.

\par 
Inspired by the above update, we propose an online variant of Algorithm~\ref{alg: complete mini-batch}. We start the algorithm by initializing the preconditioner and the data matrix using $p_1$ and $p_2$ samples, respectively, both potentially significantly smaller than $p$. When a new sample arrives, we update the preconditioner $\bP^{(t)}$ via the triangular rank-one update and update the data set accordingly. 

\begin{algorithm}
	\caption{Alternating minimization for online dictionary learning}
	\label{alg: complete mini-batch}
	\begin{algorithmic}[1]
		\STATE{{\bf Input:} $\bA^{(0)}$, $\zeta$, $p_1$,$p_2$}
		\STATE{Set $\bZ^{(0)}_{\text{inv}} = \left(\bar \bY \bar \bY^\top\right)^{-1}$, $\bP^{(0)} = \mathcal{L}\left({p_1\theta \sigma^2}\bZ^{(0)}_{\text{inv}}\right)^\top$, where $\bar \bY$ is constructed with $p_1$ samples and $\bD^{(0)} \!=\!  \bP^{(0)}\bA^{(0)}$.}
		\STATE{Initialize $\bY^{(0)}$ with $p_2$ samples.}
		\FOR{$t = 0,1,\dots, T-1$}
		\STATE{Given a new $\by$, set $\bY^{(t)}\!\!=\!\![\bY^{(t-1)}\ \by]$ and remove the first column of $\bY^{(t)}$.}
		\STATE{Update $\bP^{(t)}$ and $\bZ^{(t)}_{\text{inv}}$ using $\bP^{(t-1)}$, $\bZ^{(t-1)}_{\text{inv}}$, and $\by$ via Algorithm~\ref{alg: precon update}.}\label{step update P}
		\STATE{Set $\tbY^{(t)} = \bP^{(t)}\bY^{(t)}$}
		\STATE{Set $\tbX^{(t)} = \mathrm{HT}_{\zeta}\left(\bD^{(t)\top}\tbY^{(t)}\right)$.}
		\STATE{Set $\bD^{(t+1)} =  \mathrm{Polar}\left(\tbY^{(t)} \tbX^{(t)\top}\right)$.}
		\ENDFOR
		\RETURN{$\left(\bP^{(T-1)}\right)^{-1}\bD^{(T)}$ as an approximation to $\bA^*$}
	\end{algorithmic}
\end{algorithm}

Our next theorem establishes the convergence of Algorithm~\ref{alg: complete mini-batch}.
\begin{theorem}\label{theorem: complete mini-batch}
Suppose that $\bY = \bA^* \bX^*$, where $\bX^*$ satisfies Assumption \ref{assump:sparse and normalization} with sparsity level $0<\theta < 1/2$.
    Suppose that the initial dictionary $\bA^{(0)}$ satisfies $\|\bA^{(0)}\!-\!\bA^*\|_F \lesssim \frac{1}{\hkappa\sigma\sqrt{\theta n}}$ for some $(\bA^*, \bX^*)\in \cM$. Moreover, suppose that
    $ n/\theta^2 \lesssim p_2 \lesssim n^\gamma$, $\hkappa^{12}n^{3}\log^2 p_2/\theta\lesssim p_1$ and $T \lesssim n^\beta$, for some constants $\gamma, \beta>0$. Then, with probability at least $1-n^{-\omega(1)}$, the output of Algorithm~\ref{alg: complete mini-batch} with  $\zeta=\Gamma/2$ satisfies:
    \begin{align}
        \left\|\bA^{(T)}-\bA^*\right\|_F\le &\ 0.9^T\|\bA^{(0)}-\bA^*\|_F+O\left(\frac{n\hkappa^6}{\theta\sqrt{p_1+T}}\right).
    \end{align}
\end{theorem}
The convergence result of Algorithm~\ref{alg: complete mini-batch} is similar to that of Algorithm~\ref{alg: complete offline}, with a key difference that the statistical error $O\left(\frac{n\hkappa^6}{\theta\sqrt{p_1+T}}\right)$ now decreases with $T$, which is due to the fact that the preconditioner becomes progressively more accurate as new samples continue to arrive. 
\subsection{Initialization}\label{subsec_init}
The theoretical success of Algorithms~\ref{alg: ortho offline}-\ref{alg: complete mini-batch} requires, at least in theory, a good initialization with $O(1/\sqrt{\theta n})$ distance to the ground truth. Such initialization can be provided by the initialization scheme introduced in \cite{agarwal2016learning} and \cite{arora2015simple}, albeit with slightly more restrictive conditions on the sparsity level. For the same generative model as described in Assumption~\ref{assump:sparse and normalization}, computationally intensive and data-intensive algorithms, such as the Riemannian Trust Region method (RTR) \cite{sun2015complete}, can also be employed to design initial dictionaries that satisfy the conditions outlined in Theorems~\ref{theorem: ortho offline}-\ref{theorem: complete mini-batch}. Specifically, for ODL, \cite[Theorem 3.1]{sun2015complete} demonstrates that the RTR can obtain an initial point with a distance of $O(1/\sqrt{\theta n})$ from the ground truth, albeit with an increased sample complexity of $p = \Omega(n^5)$. Similarly, for CDL, \cite[Theorem 3.2]{sun2015complete} illustrates that RTR can yield an initial point with a distance of $O(1/\sqrt{\theta n})$ from the ground truth with a sample complexity of $p = \Omega(n^7)$. Combining these results with Theorems~\ref{theorem: ortho offline}-\ref{theorem: complete mini-batch} allows for an end-to-end global guarantee at the cost of increasing the sample complexity to $\Omega(n^5)$ for ODL and $\Omega(n^7)$ for CDL.

However, we note that these tailored initialization schemes are not easily implementable, especially in a huge-scale regime. In practice, we observed that a simple warm-start algorithm would yield similar performance. Consider a variant of Algorithm~\ref{alg: ortho offline} with a diminishing threshold presented in Algorithm~\ref{alg: initialization} (the warm-start schemes for Algorithms~\ref{alg: complete offline} and~\ref{alg: complete mini-batch} are similar and hence omitted for brevity).

\begin{figure}
    \centering
        \includegraphics[scale=0.22]{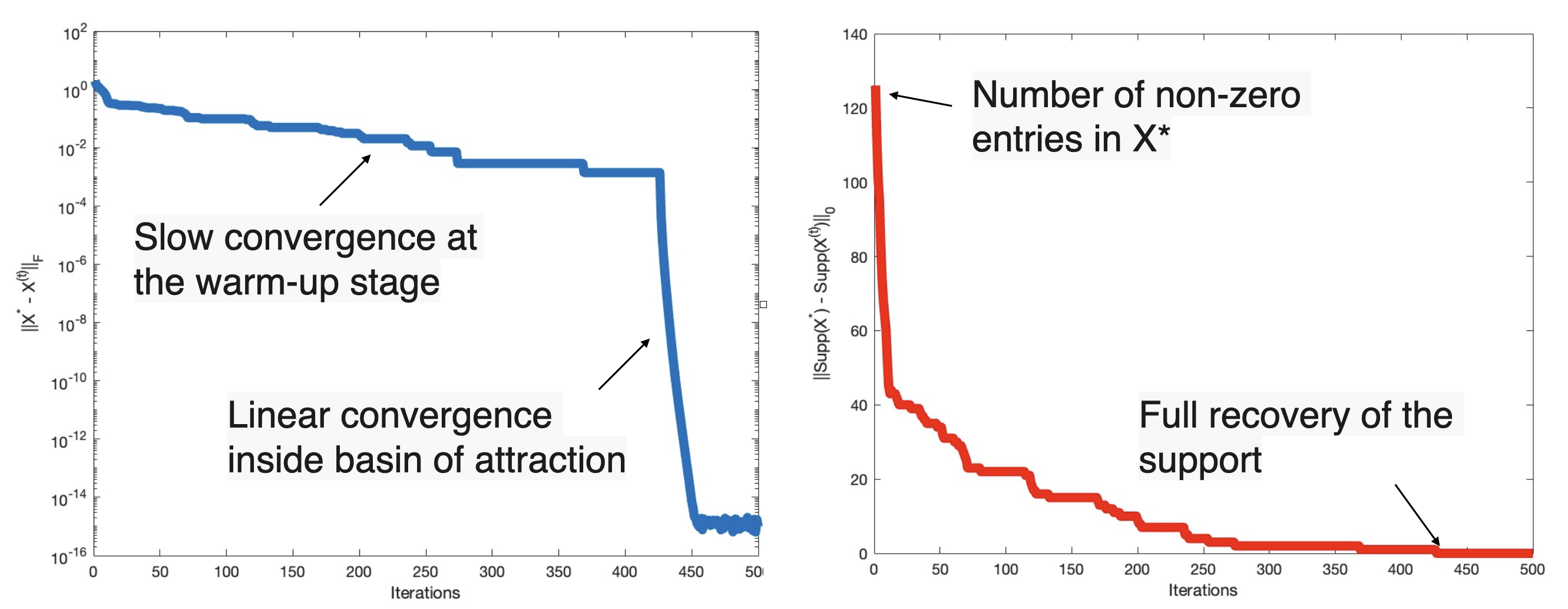}
        \caption{The plots above show the iterates of Algorithm~\ref{alg: initialization} with $p=100$, $n=5$ and $\theta=0.3$. The left figure shows the error in the sparse code. The right figure shows the number of non-zero entries of $\mathrm{Supp}(\bX^*)-\mathrm{Supp}(\bX^{(t)})$. The number is 126 at the beginning, which is the total number of non-zero entries in $\bX^*$, and 0 in the end, which indicates the full recovery of the support of $\bX^*$. }
        \label{fig:supp recover}
    \end{figure}

Notice that a large $\zeta_0$ will force $\bX^{(t)}$ to be an all-zero matrix and make $\bD^{(t+1)}$ an identity matrix. As we shrink the threshold $\zeta_t$, the matrix $\bX^{(t)}$ eventually becomes nonzero, and the iterates $\bD^{(0)}$ start to gradually move towards the basin of attraction of $\bD^*$.  Figure~\ref{fig:supp recover} demonstrates how our proposed warm-start algorithm generates an initial dictionary that is sufficiently close to the true dictionary. We also observe that, in practice, the convergence of iterates from $\bI_d$ to the basin of local convergence is not linear. However, once the iterates reach this basin, they exhibit linear convergence to the ground truth, which is theoretically guaranteed by Theorem~\ref{theorem: ortho offline}. We also note that our proposed warm-start algorithm can serve as an effective initialization scheme for other DL algorithms, including those discussed in Section~\ref{sec: related work}. The theoretical analysis of this warm-start algorithm is left as an enticing challenge for future research. 

\begin{algorithm}[H]
	\caption{Algorithm~\ref{alg: ortho offline} with warm-start}
	\label{alg: initialization}
	\begin{algorithmic}[1]
		\STATE{{\bf Input:} $\bY$, $\zeta_0$, $\zeta$, $\beta\in(0,1)$}
		\STATE{ Set $\bD^{(0)} = \bI_n$}
		\FOR{$t = 0,1,\dots, T_0$}
		\STATE{$\zeta_{t+1} =\max\{\beta \zeta_t, \zeta\}$}
		\STATE{Set $\bX^{(t)} = \mathrm{HT}_{\zeta_t}\left(\bD^{(t)\top}\bY\right)$}
		\STATE{Set $\bD^{(t+1)} = \bI_n$ if $\bX^{(t)} = \mathbf{0}_n$, and $\bD^{(t+1)} = \mathrm{Polar}\left(\bY \bX^{(t)\top}\right)$ if $\bX^{(t)} \not= \mathbf{0}_n$.}
		\ENDFOR
		\RETURN{$\bD^{(T_0)}$}
	\end{algorithmic}
\end{algorithm}

\section{Proofs}
In this section, we present the proofs of our main theorems.
 
\subsection{Proof of Theorem~\ref{theorem: ortho offline}}
We use induction to prove Theorem~\ref{theorem: ortho offline}. We consider the following induction hypothesis:
\begin{equation}\label{eq: induction hypothesis for ortho offline}
    \|\bD^{(s)}-\bD^*\|_F= O\left(\frac{1}{\sigma\sqrt{\theta n}}\right) \qquad \text{at iteration $s$.}
\end{equation}
As will be shown later, this induction hypothesis will lead to the desired linear convergence result.
The base case is verified, given the initial condition. We next present the following two key lemmas, the proofs of which can be found in Appendix~\ref{app_lemmas}.
\begin{lemma}[Exact support recovery]\label{lemma: exact support recovery}
    Consider $\bY = \bD^*\bX^*$ where $\bD^*\in\mathbb{O}^{n}$ and $\bX^*$ satisfies Assumption~\ref{assump:sparse and normalization}. For all $\bD$ satisfying $\|\bD-\bD^*\|_{1,2}= O\left(\frac{1}{\sigma\sqrt{\theta n}}\right)$, with probability at least $1-2\exp\left(\log p-Cn\right)$,  we have $\supp\left(\mathrm{HT}_{\Gamma/2} \left(\bD^{\top}\bY\right)\right) = \supp\left(\bX^*\right)$.
\end{lemma}

\begin{lemma}[Guaranteed improvement on polar decomposition]\label{lemma: improvement on polar}
Consider $\bY = \bD^*\bX^*$ where $\bD^*\in\mathbb{O}^{n}$ and $\bX^*$ satisfies Assumption~\ref{assump:sparse and normalization} with sparsity level $0<\theta<1/2$ and suppose that $p= \Omega\left( n/\theta^2\right)$. Then, with probability at least $1-2\exp(\log n-n)$, the following inequality holds for every approximation $\bX$ of $\bX^*$ such that $\frac{\|\bX-\bX^*\|_F}{\|\bX^*\|_2}= O(1)$ and $\supp(\bX) = \supp(\bX^*)$:
	\begin{equation}\nonumber
	    \|\mathrm{Polar}(\bY \bX^{\top})-\bD^{*}\|_F< 0.9\frac{\|\bX-\bX^*\|_F}{\|\bX^*\|_2}.
	\end{equation}
\end{lemma}
Lemma~\ref{lemma: exact support recovery} together with our induction hypothesis~\eqref{eq: induction hypothesis for ortho offline} implies that
\begin{equation}\label{eq: supp equivalency ortho offline}
    \supp\left(\bX^{(s)}\right) = \supp\left(\mathrm{HT}_{\Gamma/2}(\bD^{(s)\top}\bY)\right) = \supp\left(\bX^*\right).
\end{equation}
We can use the exact recovery of support to bound the error on the sparse code:
\begin{align}\nonumber
    \|\bX^{(s)}-\bX^*\|_F &= \|\mathrm{HT}_{\Gamma/2}(\bD^{(s)\top}\bY)-\bX^*\|_F\nonumber \overset{(a)}{\le} \|\bD^{(s)}-\bD^*\|_F\|\bY\|_2,
\end{align}
where we used~\eqref{eq: supp equivalency ortho offline} for (a).
The above inequality implies that
\begin{align}\label{eq: bound on sparse code ortho offline}
    \frac{\|\bX^{(s)}-\bX^*\|_F}{\|\bX^*\|_2} 
    &= \frac{\|\bX^{(s)}-\bX^*\|_F}{\|\bD^*\bX^*\|_2}= \frac{\|\bX^{(s)}-\bX^*\|_F}{\|\bY\|_2}\le \|\bD^{(s)}-\bD^*\|_F= O\left(\frac{1}{\sigma\sqrt{\theta n}}\right)
\end{align}
Invoking Lemma~\ref{lemma: improvement on polar} with $\bX = \bX^{(s)}$ together with~\eqref{eq: bound on sparse code ortho offline} leads to
\begin{equation}\nonumber
    \|\bD^{(s+1)}-\bD^{*}\|_F< \frac{0.9\|\bX^{(s)}-\bX^*\|_F}{\|\bX^*\|_2}< 0.9\|\bD^{(s)}-\bD^*\|_F.
\end{equation}
Therefore, the induction hypothesis~\eqref{eq: induction hypothesis for ortho offline} holds for $t=s+1$. Consequently, the linear convergence for $\frac{\|\bX^{(s)}-\bX^*\|_F}{\|\bX^*\|_2}$ follows from~\eqref{eq: bound on sparse code ortho offline} and the above inequality:
\begin{align*}
    \frac{\|\bX^{(s+1)}-\bX^*\|_F}{\|\bX^*\|_2}&\le \|\bD^{(s+1)}-\bD^{*}\|_F< 0.9\frac{\|\bX^{(s)}-\bX^*\|_F}{\|\bX^*\|_2}.
\end{align*}
The proof is complete by noticing a lower bound of the probability that both Lemma~\ref{lemma: exact support recovery} and Lemma~\ref{lemma: improvement on polar} hold is $1-2\exp\left(\log p-Cn\right)-2\exp(\log n-n)=1-n^{-\omega(1)}$, which follows from our assumption $p = O(n^\gamma)$. This completes the proof.$\hfill\square$

Next, we provide the proof of Theorem~\ref{theorem: complete mini-batch}. As will be shown later, the proof of Theorem~\ref{theorem: complete offline} readily follows from this proof. 
\subsection{Proof of Theorem~\ref{theorem: complete mini-batch}}
The linear convergence of $\bA^{(T)}$ hinges on the linear convergence of the preconditioned dictionary $\bD^{(T)}$ towards
\begin{equation}\nonumber
    \bD^* = \mathcal{L}((\bA^*\bA^{*\top})^{-1})^\top\bA^*.
\end{equation}
To establish this fact, we first prove the following inequality:
\begin{align}\label{eq: induction step for complete offline}
    \|\bD^{(s+1)}-\bD^*\|_F\le 0.9\|\bD^{(s)}-\bD^*\|_F + O\left(\|\bP^{(s)}\bA^*-\bD^*\|_F\right),\quad s=0,2,\dots, T-1
\end{align}
We present the following intermediate lemmas, the proofs of which are in Appendix~\ref{app_lemmas}.
 \begin{lemma}[Bounding preconditioner error]\label{lemma: tilde Y to real Y}
Suppose that $\bX^*$ satisfies Assumption \ref{assump:sparse and normalization}. Then, with probability at least $1-2\exp(-n)$, we have:
\begin{equation}\nonumber
    \left\|\bP^{(t)}-\mathcal{L}\left(\left(\bA^*\bA^{*\top}\right)^{-1}\right)^\top\right\|_2 \le O\left(\frac{\hkappa^6}{\theta}\sqrt{\frac{n}{p_1+t}}\right).
\end{equation}
\end{lemma}
\begin{lemma}[Spectral property for sparse code matrix]\label{lemma: covariance estimation on X}
Suppose that $\bX^*$ satisfies Assumption~\ref{assump:sparse and normalization} and $p= \Omega(\tau n/\theta^2)$ for an arbitrary $\tau>0$. Then, with probability at least $1-2\exp(-n)$, we have 
\begin{align}\nonumber
    \|\bX^*\|_2\le \left(1+\tau^{-\frac{1}{4}}\right)&\sqrt{p\theta}\sigma,\quad
    \sigma_n\left(\bX^*\right)\ge \left(1-\tau^{-\frac{1}{4}}\right)\sqrt{p\theta}\sigma,\quad\kappa\left(\bX^*\right)\le \frac{1+\tau^{-\frac{1}{4}}}{1-\tau^{-\frac{1}{4}}}.
\end{align}
\end{lemma}
 Similar with the proof of Theorem~\ref{theorem: complete offline}, we will use the induction hypothesis $\|\bD^{(s)}-\bD^*\|_F\le O\left(\frac{1}{\sigma\sqrt{\theta n}}\right)$. The base case is easy to verify given the initial condition and Lemma~\ref{lemma: tilde Y to real Y}:
 \begin{align}
     &\quad\|\bD^{(0)}-\bD^*\|_F = \|\bP^{(0)}\bA^{(0)}-\mathcal{L}((\bA^*\bA^{*\top})^{-1})^\top\bA^*\|_F\nonumber\\
     &= \|\bP^{(0)}\bA^{(0)} - \mathcal{L}((\bA^*\bA^{*\top})^{-1})^\top\bA^{(0)}+\mathcal{L}((\bA^*\bA^{*\top})^{-1})^\top\bA^{(0)}-\mathcal{L}((\bA^*\bA^{*\top})^{-1})^\top\bA^*\|_F\nonumber\\
     &\le \|\bP^{(0)}-\mathcal{L}((\bA^*\bA^{*\top})^{-1})^\top\|_2\|\bA^{(0)}\|_F+\|\mathcal{L}((\bA^*\bA^{*\top})^{-1})^\top\|_2\|\bA^{(0)}-\bA^*\|_F\nonumber \\
     &\le O\left(\frac{\hkappa^6}{\theta}\sqrt{\frac{n}{p_1}} \cdot \sqrt{n}\right) + O\left(\hkappa \cdot \frac{1}{\hkappa\sigma\sqrt{\theta n}} \right) = O\left(\frac{1}{\sigma\sqrt{\theta n}}\right).\nonumber
 \end{align}
 Moreover, after achieving~\eqref{eq: induction step for complete offline} for each $s$, we have $ \|\bD^{(s+1)}-\bD^*\|_F=O\left(\frac{1}{\sigma\sqrt{\theta n}}\right)$ since
 \begin{equation*}
     \|\bP^{(s)}\bA^*-\bD^*\|_F\le \left\|\bP^{(s)}-\mathcal{L}\left(\left(\bA^*\bA^{*\top}\right)^{-1}\right)^\top\right\|_2\|\bD^*\|_F=O\left(\frac{1}{\sigma\sqrt{\theta n}}\right).
 \end{equation*}
 For the remainder of this section, we abuse the notation and use $\bX^*$ to denote the sparse code that generates $\bY^{(s)}=\bA^*\bX^*$, where $\bY^{(s)}$ is obtained by adding the latest sample as the last column and removing the oldest sample from its first column (see Steps 6 and 7 in Algorithm~\ref{alg: complete mini-batch}). Indeed, $\bX^*$ is different from iteration to iteration, but it plays a similar role in the proof to the ground truth matrix in the full-batch case. We also define $\tbX$ as $\tbX =\bD^{*\top}\tbY^{(s)}$, which is again a different sparse code for different iterations since $\tbY^{(s)}=\bP^{(s)}\bY^{(s)}$ and both $\bP^{(s)}$ and $\bY^{(s)}$ change from iteration to iteration. Let us assume that the initial sample size and batch size satisfy $p_1= \Omega(\tau_1n^3\log ^2 p_2\sigma^2\hkappa^{12}/(\theta \Gamma^2))$ and $p_2=\Omega(\tau^4_{2}n/\theta^2)$, for parameters $\tau_1,\tau_2>0$ to be defined later.\par
 
Towards proving~\eqref{eq: induction step for complete offline}, we first show that $\tbX^{(s)}$ recovers the sparsity pattern of $\bX^*$ if we use enough number of samples to construct the preconditioner. To show this, we consider one entry $\left( \bD^{(s)\top}\tbY^{(s)}\right)_{ij}$ of $\bD^{(s)\top}\tbY^{(s)}$ and write
\begin{align}\label{eq: decompose of complete entry}
    \left( \bD^{(s)\top}\tbY^{(s)}\right)_{ij}
    =\underbrace{\left( \bD^{(s)\top}\bD^*(\tbX-\bX^*)\right)_{ij}}_{:=\cA_{ij}}+\underbrace{\left( \bD^{(s)\top}\bD^*\bX^*\right)_{ij}}_{:=\cB_{ij}}.
\end{align}
The first term in the right-hand side can be bounded as
\begin{align}\label{eq_DDXX}
    & \left |\cA_{ij}\right|
    \le \  \| \bD^{(s)\top}\bD^*\|_2\|\tbX-\bX^*\|_{1,2}
     = \|\tbX-\bX^*\|_{1,2}\nonumber\\
    & \le \|\bD^{*\top}\bP^{(s)}\bA^*\bX^*-\bX^*\|_{1,2}
    \le \|\bP^{(s)}\bA^*-\bD^*\|_2\|\bX^*\|_{1,2}\nonumber\\
    &\le \left\|\bP^{(s)}-\mathcal{L}\left(\left(\bA^*\bA^{*\top}\right)^{-1}\right)\right\|_2\|\bA^*\|_2\|\bX^*\|_{1,2}
    \lesssim \frac{\hkappa^6}{\theta}\sqrt{\frac{n}{p_1+T}}\|\bX^*\|_{1,2}.
\end{align}
with probability at least $1-2\exp(- n)$, where the last inequality is due to Lemma~\ref{lemma: tilde Y to real Y}. The expectation for the norm of each column, based on Assumption~\ref{assump:sparse and normalization}, is $\sigma\sqrt{\theta n}$. Therefore, according to Theorem~\ref{thm: norm concentration} (see Appendix~\ref{app_pre}), the random variable $\|\bX^*_{(\cdot,k)}\|_2-\sigma\sqrt{\theta n}$ is sub-Gaussian with sub-Gaussian norm $O(\sqrt{\theta}\sigma)$. As a result, the $k$th column of $\bX^*$ satisfies
\begin{equation}\label{eq: concentration for each column of X}
    \mathbb{P}\left(\|\bX^*_{(\cdot,k)}\|_2\ge 2\sigma\sqrt{\theta n}\right)\le 2\exp\left(-Cn\right),
\end{equation}
for some constant $C$. Upon taking a union bound over different columns, we have 
\begin{equation}\label{eq: 2 inf norm bound complete offline}
    \|\bX^*\|_{1,2} = \max_{1\le k \le p_2}\|\bX_{(\cdot,k)}^*\|_2 \le 2\sigma\sqrt{\theta n\log p_2}
\end{equation}
with probability at least $1-2\exp\left(-Cn\right)$. Combining this inequality with~\eqref{eq_DDXX}, we have 
\begin{equation}\nonumber
    \left|\cA_{ij}\right|\lesssim  \frac{\hkappa^6\sigma}{\sqrt{\theta}}\sqrt{\frac{n^2\log p_2}{p_1+T}},
\end{equation}
for every $(i,j)$. Based on the above inequality and with the choice of $p_1 \gtrsim \frac{n^2\log p_2\sigma^2\hkappa^{12}}{\theta \Gamma^2}$, we have $ \left|\cA_{ij}\right|<\Gamma/4$. Following the proof of Lemma~\ref{lemma: exact support recovery}, the deviation of the second term $\cB_{ij}$ from $\bX^*_{ij}$ can also be upper bounded by $\Gamma/4$ with high probability, which will not be repeated here. Therefore, we have $\left|\left( \bD^{(s)\top}\tbY^{(s)}\right)_{ij}\right|<\Gamma/2$ when $\bX^*_{ij}=0$ and $\left|\left( \bD^{(s)\top}\tbY^{(s)}\right)_{ij}\right|>\Gamma/2$ when $\bX^*_{ij}\ne 0$, which leads to $\supp\left(\tbX^{(s)}\right) = \supp(\bX^*)$ with probability at least $1-2\exp(-n)-2\exp(-Cn)-2\exp\left(\log p_2-Cn\right)$. For any matrix $\bM\in\mathbb{R}^{n\times p_2}$, we define the projection operator $\mathcal{P}$ as 
\begin{equation}\nonumber
    \left(\mathcal{P}(\bM)\right)_{ij}=
    \begin{cases}
    \bM_{ij}&\text{if}\quad(i,j)\in\supp(\bX^*)\\
    0&\text{if}\quad(i,j)\not\in\supp(\bX^*)
    \end{cases}.
\end{equation}
Given the exact support recovery, we have 
\begin{align}\nonumber
    &\|\tbX^{(s)}-\mathcal{P}(\tbX)\|_F
    =\|\mathcal{P}(\bD^{(s)\top}\tbY^{(s)})-\mathcal{P}(\tbX)\|_F\le \|\bD^{(s)\top}\tbY^{(s)}-\tbX\|_F  \\
    &= \|\bD^{(s)\top}\tbY^{(s)}-\bD^{*\top}\tbY^{(s)}\|_F \le \|\bD^{(s)}-\bD^*\|_F\|\tbY^{(s)}\|_2 = \|\bD^{(s)}-\bD^*\|_F\|\tbX\|_2.\nonumber
\end{align}
As a result we have
\begin{align}\label{eq: sparse code bounded by dic complete offline}
    &\frac{\|\tbX^{(s)}-\mathcal{P}(\tbX)\|_F}{\|\mathcal{P}(\tbX)\|_2}
    =\ \frac{\|\tbX^{(s)}-\mathcal{P}(\tbX)\|_F}{\|\tbX\|_2}\frac{\|\tbX\|_2}{\|\mathcal{P}(\tbX)\|_2} \le \|\bD^{(s)}-\bD^*\|_F\frac{\|\tbX\|_2}{\|\tbX\|_2-\|\tbX-\mathcal{P}(\tbX)\|_2} \\
    &\le \|\bD^{(s)}-\bD^*\|_F\frac{\|\tbX\|_2}{\|\tbX\|_2-\|\tbX-\mathcal{P}(\tbX)\|_F}
    \overset{(a)}{\le} \|\bD^{(s)}-\bD^*\|_F\frac{\|\tbX\|_2}{\|\tbX\|_2-\|\tbX-\bX^*\|_F}\nonumber \\
    &\overset{(b)}{\le} \|\bD^{(s)}-\bD^*\|_F\left(1+\frac{\|\tbX-\bX^*\|_F}{\|\bX^*\|_2-2\|\tbX-\bX^*\|_F}\right)
    \le \|\bD^{(s)}-\bD^*\|_F\left(1+\frac{\frac{\|\tbX-\bX^*\|_F}{\|\bX^*\|_2}}{1-\frac{2\|\tbX-\bX^*\|_F}{\|\bX^*\|_2}}\right)\nonumber \\
    & \overset{(c)}{\le} \|\bD^{(s)}-\bD^*\|_F\left(1+\frac{\frac{1}{\tau_1}}{1-2\times\frac{1}{\tau_1}}\right)
    =\ \frac{\tau_1-1}{\tau_1-2}\|\bD^{(s)}-\bD^*\|_F.\nonumber
\end{align}
Here inequality (a) is due to the decomposition of $\tbX-\bX^*$ onto and outside the support of $\bX^*$. Specifically, define $\mathcal{P}_\perp(\cdot)$ as the projection onto the complement of $\mathcal{P}(\cdot)$. We have $\mathcal{P}_\perp(\tbX-\mathcal{P}(\tbX))=\mathcal{P}_\perp(\tbX-\bX^*)$ and $\mathcal{P}(\tbX-\mathcal{P}(\tbX)) = 0$, which in turn implies $\|\tbX-\mathcal{P}(\tbX)\|_F\le\|\tbX-\bX^*\|_F$. The inequality (b) is due to $\|\bX^*\|_2-\|\tbX-\bX^*\|_F\le\|\bX^*\|_2-\|\tbX-\bX^*\|_2\le \|\tbX\|_2$. The inequality (c) holds due to $\frac{\|\tbX-\bX^*\|_F}{\|\bX^*\|_2}\le 1/\tau_1$, which is a result of Lemma~\ref{lemma: tilde Y to real Y}:
\begin{align}\label{eq: Cp1}
    \|\tbX-\bX^*\|_F&=\|\bD^{*\top}\bP^{(s)}\bA^*\bX^*-\bX^*\|_F\le \|\bP^{(s)}\bA^*-\bD^*\|_F\|\bX^*\|_2 \\
    &\le \sqrt{n}\|\bP^{(s)}\bA^*-\bD^*\|_2\|\bX^*\|_2\lesssim\frac{\hkappa^6}{\theta}\sqrt{\frac{n^2}{p_1+T}}\|\bX^*\|_2,\nonumber
\end{align}
and $p_1= \Omega(\tau_1n^3\log ^2 p_2\sigma^2\hkappa^{12}/(\theta \Gamma^2))$.\par 
So far, we have upper bounded $\frac{\|\tbX^{(s)}-\mathcal{P}(\tbX)\|_F}{\|\mathcal{P}(\tbX)\|_2}$ with $\|\bD^{(s)}-\bD^*\|_F$. Next, we provide an upper bound on $\|\bD^{(s+1)}-\bD^*\|_F$ in terms of $\frac{\|\tbX^{(s)}-\mathcal{P}(\tbX)\|_F}{\|\mathcal{P}(\tbX)\|_2}$. We have
\begin{align}\label{eq: decomp complete offline}
    &\|\bD^{(s+1)}-\bD^*\|_F 
    =\|\mathrm{Polar}(\tbY^{(s)}\tbX^{(s)\top})-\bD^*\|_F
    =\ \|\mathrm{Polar}(\bD^*\tbX\tbX^{(s)\top})-\bD^*\|_F \\
    &=\|\mathrm{Polar}(\bD^*\tbX\tbX^{(s)\top})-\mathrm{Polar}(\bD^*\mathcal{P}(\tbX)\tbX^{(s)\top})+\mathrm{Polar}(\bD^*\mathcal{P}(\tbX)\tbX^{(s)\top})-\bD^*\|_F\nonumber \\
    &=\|\mathrm{Polar}(\bD^*\tbX\tbX^{(s)\top})\!-\!\mathrm{Polar}(\bD^*\mathcal{P}(\tbX)\tbX^{(s)\top})\|_F\!+\!\|\mathrm{Polar}(\bD^*\mathcal{P}(\tbX)\tbX^{(s)\top})\!-\!\bD^*\|_F\nonumber
\end{align}
To control the first term on the right-hand side, we invoke Theorem~\ref{thm: perturbation bound for polar decomposition} (see Appendix~\ref{app_pre}): 
\begin{align}\nonumber
    &\|\mathrm{Polar}(\bD^*\tbX\tbX^{(s)\top})-\mathrm{Polar}(\bD^*\mathcal{P}(\tbX)\tbX^{(s)\top})\|_F \\
    \le \ & \frac{2\|\bD^*\tbX\tbX^{(s)\top}-\bD^*\mathcal{P}(\tbX)\tbX^{(s)\top}\|_F}{\sigma_n(\bD^*\tbX\tbX^{(s)\top})+\sigma_n(\bD^*\mathcal{P}(\tbX)\tbX^{(s)\top})}\nonumber \\
    \overset{(a)}{\le}  \ &\frac{2\kappa(\tbX^{(s)})\|\tbX-\mathcal{P}(\tbX)\|_F}{2\sigma_n(\bX^*)-\|\bX^*-\tbX\|_2-\|\bX^*-\mathcal{P}(\tbX)\|_2} \nonumber \\
    = \ &\frac{2\kappa(\tbX^{(s)})\|\tbX-\mathcal{P}(\tbX)\|_F}{\|\bX^*\|_2}\cdot\frac{1}{\frac{2}{\kappa(\bX^*)}-\frac{\|\bX^*-\tbX\|_2}{\|\bX^*\|_2}-\frac{\|\bX^*-\mathcal{P}(\tbX)\|_2}{\|\bX^*\|_2}} \nonumber \\
    \overset{(b)}{\le} \ & \frac{\tau_1(\tau_2-1)}{\tau_1(\tau_2+1)-\tau_2+1}\frac{\kappa(\tbX^{(s)})\|\tbX-\mathcal{P}(\tbX)\|_F}{\|\bX^*\|_2} \nonumber
\end{align}
where (a) follows after normalizing both the numerator and the denominator with $\|\tbX^{(s)}\|_2$. Moreover, (b) follows from~\eqref{eq: Cp1}, Lemma~\ref{lemma: covariance estimation on X} and $p_2=\Omega(\tau^4_2n/\theta^2)$. Next, we provide a bound on $\kappa(\tbX^{(s)})$, and we do so by first bounding $\kappa(\mathcal{P}(\tbX))$. By Lemma~\ref{lemma: covariance estimation on X} and Lemma~\ref{lemma: tilde Y to real Y}, we have 
\begin{align}\label{eq: kappa P tilde X complete offline}
    &\kappa(\mathcal{P}(\tbX))
    =\frac{\|\mathcal{P}(\tbX)\|_2}{\sigma_n(\mathcal{P}(\tbX))}\le\frac{\|\bX^*\|_2+\|\mathcal{P}(\tbX)-\bX^*\|_2}{\sigma_n(\bX^*)-\|\mathcal{P}(\tbX)-\bX^*\|_2}\\
    &=\frac{\|\bX^*\|_2+\|\mathcal{P}(\tbX)-\bX^*\|_2}{\|\bX^*\|_2/\kappa(\bX^*)-\|\mathcal{P}(\tbX)-\bX^*\|_2}\le
    \frac{\tau_1+1}{\frac{\tau_1(\tau_2-1)}{\tau_2+1}-1}\nonumber
\end{align}
Upon assuming $\tau_1\ge 10$ and $\tau_2\ge 10$, we have
\begin{align}\nonumber
    \kappa(\tbX^{(s)}) &\!\le\!\frac{\|\mathcal{P}(\tbX)\|_2\!+\!\|\mathcal{P}(\tbX)\!-\!\tbX^{(s)}\|_2}{\sigma_n(\mathcal{P}(\tbX))\!-\!\|\mathcal{P}(\tbX)\!-\!\tbX^{(s)}\|_2}\!\le\!\frac{1+\|\mathcal{P}(\tbX)-\tbX^{(s)}\|_F/\|\mathcal{P}(\tbX)\|_2}{1/\kappa(\mathcal{P}(\tbX))-\|\mathcal{P}(\tbX)-\tbX^{(s)}\|_F/\|\mathcal{P}(\tbX)\|_2}\\
    &\overset{(a)}{\le}\frac{1+9/32}{79/121-9/32}<4.\nonumber
\end{align}
For (a), we used~\eqref{eq: sparse code bounded by dic complete offline} and the fact that $\|\bD^{(s)}-\bD^*\|_F\le 1/4$. To conclude, we have
\begin{align}
    \|\mathrm{Polar}(\bD^*\tbX\tbX^{(s)\top})-\mathrm{Polar}(\bD^*\mathcal{P}(\tbX)\tbX^{(s)\top})\|_F
    &<\frac{12\|\tbX-\mathcal{P}(\tbX)\|_F}{\|\bX^*\|_2}\le\frac{12\|\bX^*-\mathcal{P}(\tbX)\|_F}{\|\bX^*\|_2}.\nonumber
\end{align}
Next, we turn our focus to $\|\mathrm{Polar}(\bD^*\mathcal{P}(\tbX)\tbX^{(s)\top}) -\bD^*\|_F$, which is the second term of~\eqref{eq: decomp complete offline}. We first introduce the following lemma, which is borrowed from \cite{ravishankar2020analysis}:
\begin{lemma}[Approximation error for dictionary (See the proof of Lemma 2.4 in \cite{ravishankar2020analysis})]\label{q_lm}
    For any matrix $\bX^*\in\mathbb{R}^{n\times p}$ and $\bY=\bD^*\bX^*$ where $\bD^*\in\mathbb{O}^{n}$, consider an approximation $\bX$ to $\bX^*$ with $\bE:=\left(\bX-\bX^*\right)/\|\bX^*\|_2$ satisfying $\|\bE\|_F<{1}/{\kappa^2\left(\bX^*\right)}$.
    Then, for some constant $C_T>0$, $\bD = \mathrm{Polar}(\bY\bX^\top)$ satisfies:
    \begin{align}\nonumber
        \|\bD - \bD^*\|_F
    \le&\frac{\kappa(\bX^*)^4}{2}\frac{\left\|\bX^{*}\bE^\top-\bE \bX^{*\top}\right\|_F}{\|\bX^*\|_2}+C_T\|\bE\|^2_F.
    \end{align}
\end{lemma}
To invoke Lemma~\ref{q_lm}, let $\tbE^{(s)} = \frac{\tbX^{(s)}-\mathcal{P}(\tbX)}{\|\mathcal{P}(\tbX)\|_2}$, and notice that 
\begin{align}\nonumber
    \|\tbE^{(s)}\|_F&\overset{(a)}{\le} \frac{9}{8}\|\bD^{(s)}-\bD^*\|_F
    \le \frac{9}{32}\le  \left(\frac{79}{121}\right)^2
    \overset{(b)}{\le}\frac{1}{\kappa^2(\mathcal{P}(\tbX))}.
\end{align}
where (a) and (b) follow from~\eqref{eq: sparse code bounded by dic complete offline} and \eqref{eq: kappa P tilde X complete offline}, respectively.
Therefore, Lemma~\ref{q_lm} can be invoked:
\begin{align}\label{eq: invoke of q_lm complete offline}
    \|\mathrm{Polar}(\bD^*\mathcal{P}(\tbX)\tbX^{(s)\top}) -& \bD^*\|_F\le
\frac{\kappa(\mathcal{P}(\tbX))^4}{2}\frac{\left\|\mathcal{P}(\tbX)\tbE^{(s)\top}-\tbE^{(s)} \mathcal{P}(\tbX)^\top\right\|_F}{\|\mathcal{P}(\tbX)\|_2}+C_T\|\bE\|^2_F.
\end{align}
By recalling $\supp(\tbX^{(s)}) = \supp(\bX^*)$, we follow the same argument in the proof of Lemma~\ref{lemma: improvement on polar} and conclude that
\begin{align}\nonumber
    &\frac{\left\|\mathcal{P}(\tbX)\tbE^{(s)\top}-\tbE^{(s)} \mathcal{P}(\tbX)^\top\right\|_F}{2\|\mathcal{P}(\tbX)\|_2}\le \frac{\max_{1\le k \le n} \left\|\bM_{k}\mathcal{P}(\tbX)\tilde{\bM}_{k}\right\|_2\|\tbE^{(s)} \|_F}{\|\mathcal{P}(\tbX)\|_2},
\end{align}
where $\bM_{k}$ and $\tilde{\bM}_{k}$ are defined in the same way as in the proof of Lemma~\ref{lemma: improvement on polar}. Since both $\bM_{k}$ and $\tilde{\bM}_{k}$ are linear operators, we have
\begin{align}
    & \frac{\max_{1\le k \le n} \left\|\bM_{k}\mathcal{P}(\tbX)\tilde{\bM}_{k}\right\|_2}{\|\mathcal{P}(\tbX)\|_2}
    \le \frac{\max_{1\le k \le n} \left\|\bM_{k}\bX^*\tilde{\bM}_{k}\right\|_2+\|\mathcal{P}(\tbX)-\bX^*\|_2}{\|\bX^*\|_2(1-\|\mathcal{P}(\tbX)-\bX^*\|_F/\|\bX^*\|_2)}\nonumber\\
    \le &\ \frac{\tau_1}{\tau_1-1}\!\left(\frac{\max_{1\le k \le n} \left\|\bM_{k}\bX^*\tilde{\bM}_{k}\right\|_2}{\|\bX^*\|_2}\!+\!\frac{1}{\tau_1}\right)
\end{align}
Following the proof of Lemma~\ref{lemma: improvement on polar}, we have that with probability $1-2\exp(\log n-n)$ and $p_2 = \Omega(\tau^4_2n/\theta^2)$
\begin{equation}\label{eq: MXM complete offline}
    \frac{\max_{1\le k \le n} \left\|\bM_{k}\bX^*\tilde{\bM}_{k}\right\|_2}{\|\bX^*\|_2}\le \frac{\tau_2+1}{\tau_2-1}\sqrt{\theta}.
\end{equation}
Combined with~\eqref{eq: invoke of q_lm complete offline}, we have
\begin{align}
     &\|\mathrm{Polar}(\bD^*\mathcal{P}(\tbX)\tbX^{(s)\top}) - \bD^*\|_F\le \frac{\kappa(\mathcal{P}(\tbX))^4}{2}\frac{\left\|\mathcal{P}(\tbX)\tbE^{(s)\top}-\tbE^{(s)} \mathcal{P}(\tbX)^\top\right\|_F}{\|\mathcal{P}(\tbX)\|_2}+C_T\|\tbE^{(s)}\|^2_F\nonumber \\
    &\le \left(\frac{\tau_1}{\tau_1-1}\left(\frac{\tau_2+1}{\tau_2-1}\sqrt{\theta}+\frac{1}{\tau_1}\right)\kappa(\mathcal{P}(\tbX))^4+C_T\|\tbE^{(s)}\|_F\right)\|\tbE^{(s)}\|_F\nonumber
\end{align}

Therefore, we have a bound for the second term on the right-hand side of~\eqref{eq: decomp complete offline}. Recall that$\kappa(\mathcal{P}(\tbX))\le \frac{\tau_1+1}{\frac{\tau_1(\tau_2-1)}{\tau_2+1}-1}$ due to~\eqref{eq: kappa P tilde X complete offline} and $0<\theta<1/2$. Therefore, similar to the proof of Lemma~\ref{lemma: improvement on polar}, for sufficiently large $\tau_1$ and $\tau_2$ and $\|\tbE^{(s)}\|_F\leq \frac{1}{10{C_T}}$, ~\eqref{eq: decomp complete offline} can be rewritten as

\begin{align}\label{eq: dic bounded by sparse code complete offline}
    \|\bD^{(s+1)}-\bD^*\|_F
    &\overset{(a)}{\le}  0.9\|\bD^{(s)}-\bD^*\|_F \!+\! \frac{12\|\bX^*-\mathcal{P}(\tbX)\|_F}{\|\bX^*\|_2}\nonumber \\
    &\le0.9\|\bD^{(s)}-\bD^*\|_F + 12\|\bP^{(s)}\bA^*-\bD^*\|_F,
\end{align}
 where (a) follows from \eqref{eq: sparse code bounded by dic complete offline}. This completes the proof of the induction step~\eqref{eq: induction step for complete offline}. For each induction step, our analysis holds with probability at least $1-O\left(\exp(\log p_2 - n)\right)$. After taking the union bound, the first $T$ steps of the induction hold with probability at least $1-O\left(\exp(\log T + \log p_2 - n)\right)$, which is $1-n^{-\omega(1)}$ when $T=O(n^\beta)$ and $p_2=O(n^\gamma)$. Finally, we use~\eqref{eq: dic bounded by sparse code complete offline} to complete the proof of Theorem~\ref{theorem: complete mini-batch}. To this goal, notice that
\begin{align}\label{A - D connection}
    &\|\bA^{(s+1)}-\bA^*\|_F
    =\left\|\bP^{(s+1)-1}\bD^{(s+1)}-\left(\mathcal{L}((\bA^*\bA^{*\top})^{-1})^\top\right)^{-1}\bD^*\right\|_F \\
    &\le \|\bD^{(s+1)}-\bD^*\|_F + \left\|\bP^{(s+1)-1} - \left(\mathcal{L}((\bA^*\bA^{*\top})^{-1})^\top\right)^{-1}\right\|_F\nonumber \\
    &\le 0.9\|\bD^{(s)}\!-\!\bD^*\|_F \!+\! O\left(\|\bP^{(s)}\bA^*\!-\!\bD^*\|_F\right)
    \!+\! O\left(\left\|\bP^{(s+1)}-\mathcal{L}\left(\left(\bA^*\bA^{*\top}\right)^{-1}\right)^\top\right\|_F\right)\nonumber
\end{align}
Via a similar argument, we can write
\begin{align}\label{D - A connection}
    \|\bD^{(s)}-\bD^*\|_F
    \le \ & \|\bA^{(s)}-\bA^*\|_F + O\left(\left\|\bP^{(s)}-\mathcal{L}\left(\left(\bA^*\bA^{*\top}\right)^{-1}\right)^\top\right\|_F\right).
\end{align}

Combining~\eqref{A - D connection} and~\eqref{D - A connection}, we have
\begin{align*}
    &\|\bA^{(s+1)}-\bA^*\|_F\leq 0.9 \|\bA^{(s)}-\bA^*\|_F\\
    &+O\left(\left\|\bP^{(s)}-\mathcal{L}\left(\left(\bA^*\bA^{*\top}\right)^{-1}\right)^\top\right\|_F\!+\!\|\bP^{(s)}\bA^*\!-\!\bD^*\|_F\!+\!\left\|\bP^{(s+1)}-\mathcal{L}\left(\left(\bA^*\bA^{*\top}\right)^{-1}\right)^\top\right\|_F\right)
\end{align*}

    By invoking Lemma~\ref{lemma: tilde Y to real Y}, we can bound $\|\bP^{(s)}\bA^*-\bD^*\|_F$, $\left\|\bP^{(s)}-\mathcal{L}\left(\left(\bA^*\bA^{*\top}\right)^{-1}\right)^\top\right\|_F$, and  $\left\|\bP^{(s+1)}-\mathcal{L}\left(\left(\bA^*\bA^{*\top}\right)^{-1}\right)^\top\right\|_F$ with $O\left(\frac{\hkappa^6}{\theta}\sqrt{\frac{n}{p_1+T}}\right)$. This completes the proof. $\hfill\square$

\subsection{Proof of Theorem~\ref{theorem: complete offline}}
Theorem~\ref{theorem: complete offline} is a special case of Theorem~\ref{theorem: complete mini-batch}, where we use one pre-calculated preconditioner throughout. In particular, it is easy to see that~\eqref{eq: dic bounded by sparse code complete offline} in the proof of Theorem~\ref{theorem: complete mini-batch} holds after replacing $\bP^{(s)}$ with $\bP$: 
\begin{equation}
    \|\bD^{(s+1)}-\bD^*\|_F\le 0.9\|\bD^{(s)}-\bD^*\|_F + 12\|\bP\bA^*-\bD^*\|_F,
\end{equation}
Therefore, we have 
\begin{align}
    \|\bD^{(s+1)}-\bD^*\|_F
    &\le0.9\|\bD^{(s)}-\bD^*\|_F + 12\|\bP\bA^*-\bD^*\|_F\nonumber \\
    &\le0.9^s\|\bD^{(0)}\!-\!\bD^*\|_F\!+\! \left(0.9\!+...+\!0.9^s\right)12\|\bP\bA^*-\bD^*\|_F\nonumber \\
    &=0.9^s\|\bD^{(0)}-\bD^*\|_F + O(\|\bP\bA^*-\bD^*\|_F).\nonumber
\end{align}
Given the above inequality, one can use the same argument as in~\eqref{A - D connection} and \eqref{D - A connection} to establish the linear convergence of $\bA^{(s)}$. The details are omitted for brevity.$\hfill\square$

\section{Numerical Experiments}\label{sec: numerical}
In this section, we validate our theoretical results using synthetic and real data. All experiments are performed on a MacBook Pro 2021 with the Apple M1 Pro chip and a 16GB unified memory for a serial implementation in MATLAB 2022a. The code is available at: \url{https://github.com/lianggeyuleo/CompleteDL.git}. 
\subsection{Synthetic Dataset}
We validate our theoretical results using synthetic dataset. We consider the generative model $\bY=\bA^*\bX^*$ where $\bA^*$ is a randomly generated orthogonal or full-rank dictionary. Moreover, $\bX^*$ is generated from a Gaussian-Bernoulli distribution followed by a truncation step, where the entries $\bX^*_{ij}$ with $|\bX^*_{ij}|<\Gamma = 0.3$ are replaced by $\bX^*_{ij} = \mathrm{sgn}(\bX^*_{ij})\Gamma$. Consequently, the threshold $\zeta$ in each algorithm is set to $\Gamma/2=0.15$.\par
First, we compare the efficiency of our method in solving \ref{eq:odl} to other state-of-the-art techniques. In this experiment, we vary the signal dimension $n$ from 50 to 800 and set the sample size $p=100n$, which corresponds to the linear-sample size condition in Theorem~\ref{theorem: ortho offline}. We invoke the warm-start method in Algorithm~\ref{alg: initialization} to generate the initial points. For each trial, we stop the algorithm when the consecutive iterates are close to each other ($\|\bD^{(t-1)}-\bD^{(t-1)}\|_2\le 10^{-7}$) and record the running time and final error ($\|\bD^{(T)}-\bD^*\|_F$). 
\par
\begin{figure}
    \centering
        \includegraphics[width=0.7\textwidth]{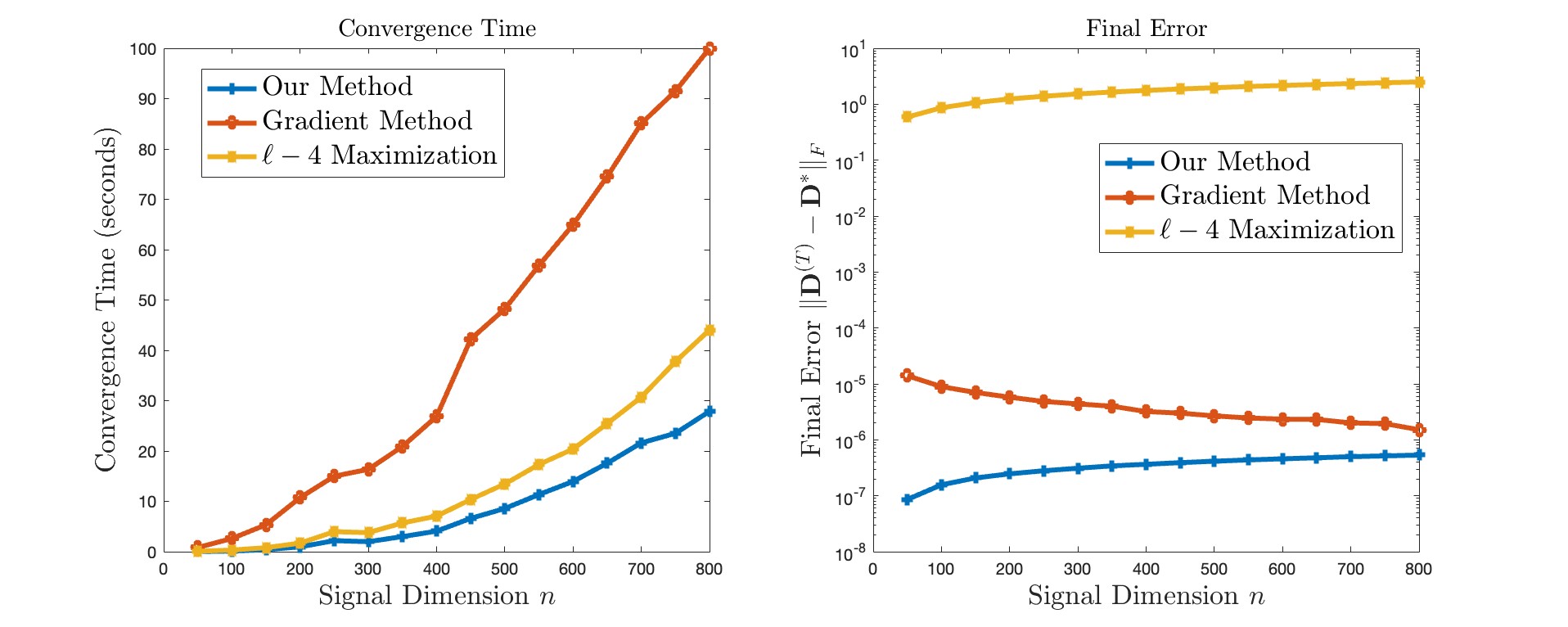}
        \caption{We compare three different dictionary learning methods with their running time and final error until convergence. 
        The results above are averaged over 5 independent trials. All methods use the same initial point. The stopping criteria is when consecutive iterates are close to each other ($\|\bD^{(t-1)}-\bD^{(t-1)}\|_2\le 10^{-7}$).   }
        \label{fig:time}
\end{figure}

Many algorithms for dictionary learning exhibit poor scalability in this setting. For instance, the convergence time of {\it KSVD method~\cite{aharon2006k}}, which is perhaps the most well-known alternating minimization algorithm for dictionary learning, exceeds 300 seconds even for our smallest instance $n=50$. Another example is the alternating minimization based on $\ell_1$-regularization~\cite{agarwal2016learning}, which suffers from a similar scalability issue. Instead, in Figure~\ref{fig:time}, we consider two other candidate algorithms that have comparable scalability with ours:
{\it Gradient-based method~\cite{arora2015simple}:} In this method, the dictionary update step (Step 4 in Algorithm~\ref{alg: ortho offline}) is replaced by a gradient step $\bD^{(t+1)} = \bD^{(t)} - \eta\nabla_{\bD^{(t)}} L$ where $L$ is the objective of \ref{eq:odl}. Here, we pick $\eta=10^{-5}$ after fine-tuning, which is the largest step-size to guarantee convergence in practice.
{\it $\ell_4$-maximization-based method~\cite{zhai2020complete}:} The work~\cite{zhai2020complete} introduces a projected gradient
ascend for maximizing the objective function $f(\bD) = \|\bD^\top\bY\|^4_4$. This method is shown to have superior performance compared to SPAMS~\cite{jenatton2010proximal} and Subgradient method~\cite{bai2018subgradient}. 
As can be seen in Figure~\ref{fig:time}, our method converges faster while achieving a smaller error. Particularly, the gradient-based method will require more time to converge, while $\ell_4$ maximization is restricted by its poor accuracy.

\begin{figure}
    \centering
        \includegraphics[width=0.7\textwidth]{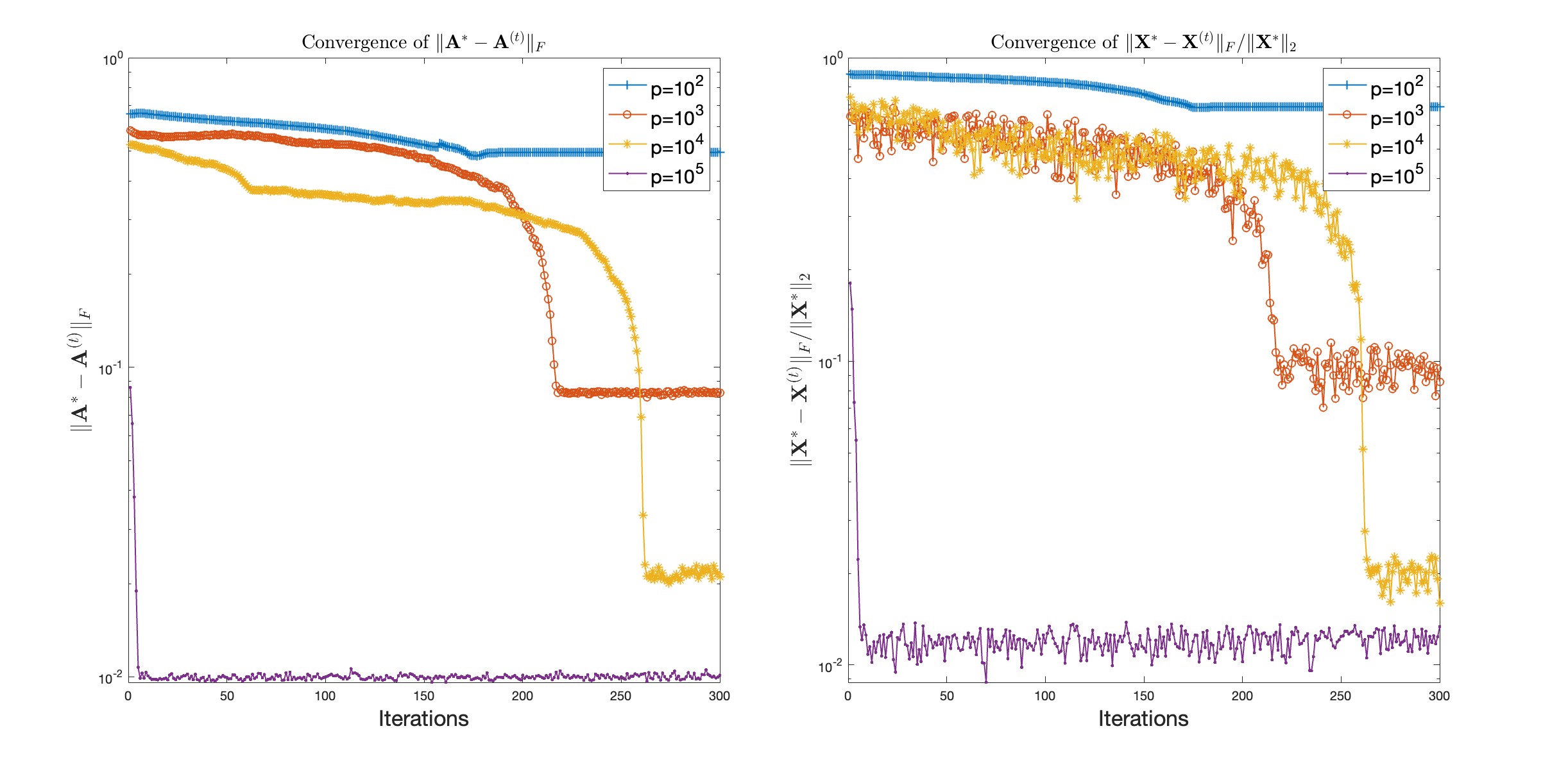}
        \caption{ Results of Algorithm~\ref{alg: complete offline} with $n=5$, $\theta = 0.3$, $\tilde{p}=100$, and varying $p$. 
        The specific number of iterations to reach convergence varies with the sample size since it depends on the distance between the ground truth and initialization.}
        \label{fig:convergence}
\end{figure}

In Figure~\ref{fig:convergence}, we report the performance of Algorithm~\ref{alg: complete offline} with warm-start for varying sample sizes. The true dictionary $\bA^*$ is a randomly generated full-rank $n\times n$ matrix, and we choose $n=5$ and $\theta = 0.3$. We set $\tilde{p}=10^2$ and vary $p$. Our experiments support our theoretical results in the following aspects: (1) we observe the fast convergence of our algorithm for different $p$, which is in line with Theorem~\ref{theorem: complete offline}. Similar to Figure~\ref{fig:supp recover}, the effect of the warm-up phase is evident in the early stages of the iterations. (2) We see clear improvement in the accuracy of the final solution with a larger sample size. Such improvement is characterized by Theorem~\ref{theorem: complete offline} with the additional error term, which diminishes to zero as $p\rightarrow \infty$.

\begin{figure}
    \centering
    \includegraphics[width=0.7\textwidth]{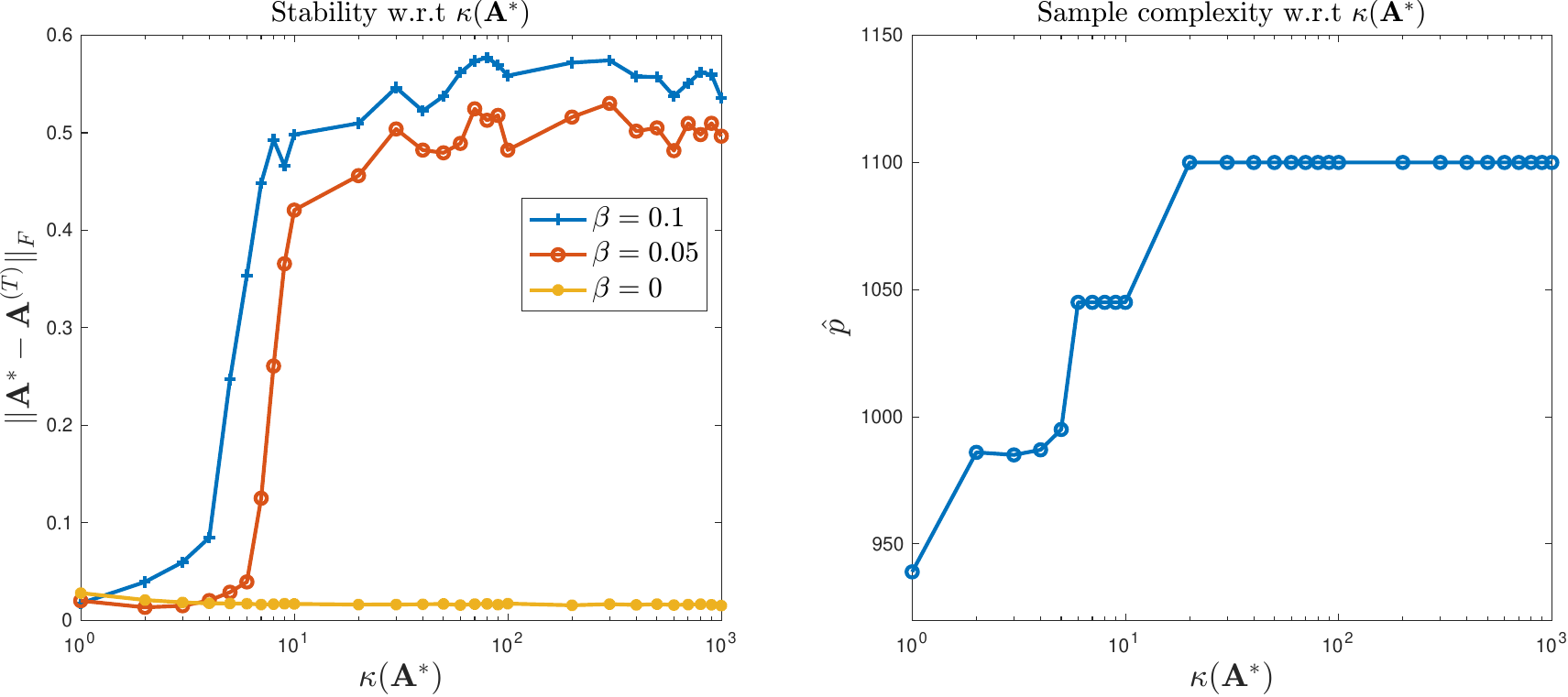}
    \caption{(left) Final error of Algorithm 2.2 in relation to condition number $\kappa(\bA^*)$ and varied noise levels $\beta$ for a fixed sample size of $\tilde{p}=p=10^5$. (right) Required sample size to achieve $\frac{\norm{\bA^{(T)}-\bA^*}_F}{\norm{\bA^*}_F}\leq 0.1$ as a function of the condition number $\kappa(\bA^*)$. In both settings, we fix $n=5$, $\theta = 0.3$, and $T=1000$.}
    \label{fig:kappa}
\end{figure}

Finally, we numerically test the stability and sample complexity of our algorithm as $\bA^*$ becomes increasingly ill-conditioned. In Figure~\ref{fig:kappa} (right), we present the results, illustrating the sample size required to achieve a final error of $\frac{\norm{\bA^{(T)}-\bA^*}_F}{\norm{\bA^*}_F}\leq 0.1$ as the condition number of $\bA^*$ varies from 1 to 1000. Surprisingly, our practical findings indicate that the relationship between sample size and the condition number differs from what our theoretical framework suggests, showcasing a more favorable scenario in practice. 

To further underscore the robustness of our algorithm, we investigate its performance in a noisy setting, where $\bY=\bA^*\bX^* + \mathbf{\mathcal{E}}$, with $\mathbf{\mathcal{E}}$ representing a noise matrix whose elements follow a Gaussian distribution with zero mean and variance $\beta^2$. As illustrated in Figure~\ref{fig:kappa} (left), our algorithm remains unaffected by the increasing condition number $\kappa(\bA^*)$ in the noiseless scenario. However, it encounters difficulties in recovering $\bA^*$ when both noise and ill-conditioning are present. We acknowledge that a theoretical explanation for this phenomenon in the noisy setting remains a topic for future investigation.


\subsection{Real Dataset}
In this section, we showcase the performance of Algorithm~\ref{alg: complete mini-batch} in learning a dictionary for the Landscape Dataset \cite{theblackmamba31_landscape_colorization} consisting of 7,000 colored images of different landscapes, 20 of which are set aside as test dataset. To gauge the quality of our learned dictionary, we use our dictionary to perform a denoising task on these images. 

\paragraph{Dataset}  Each figure in our dataset is a colored image of size $150\times 150\times 3$. Instead of directly learning a dictionary for the whole dataset, we follow the procedure in \cite{aharon2006k} and divide each figure into 225 patches of size $10\times 10 \times 3$. Each patch is then reshaped into a $300\times 1$ vector. Collecting all 225 patches over all 7000 images results in 1,575,000 patch samples. We downsample this dataset to 10,000 patch samples so that KSVD can also be applied to this dataset. This results in a data matrix $\bY$ of size $300 \times 10,000$ with $n=300$ and $p=10,000$. Our goal is to learn a dictionary $\bA$ of size $300\times 300$ for this data matrix. 

\paragraph{Image Denoising}
We gauge the quality of our learned dictionary by using it to denoise noisy images. Here, we corrupt each image with 50\% missing pixels. In other words, we select 50\% of the $150\times 150$ pixels in each image uniformly at random and then set the pixel values across all three RBG channels to $0$. Our goal is to learn a dictionary and use it to denoise this image by filling in the missing pixels. For reconstruction, we choose a sparsity level of approximately 10\%, which simply corresponds to 30 atoms for reconstruction. With a given dictionary, the reconstruction is done using a standard implementation of orthogonal matching pursuit (OMP) found in the \texttt{SPAMS} library in Matlab \cite{mairal2007sparse}. 

\begin{figure}
        \includegraphics[width=0.8\textwidth]{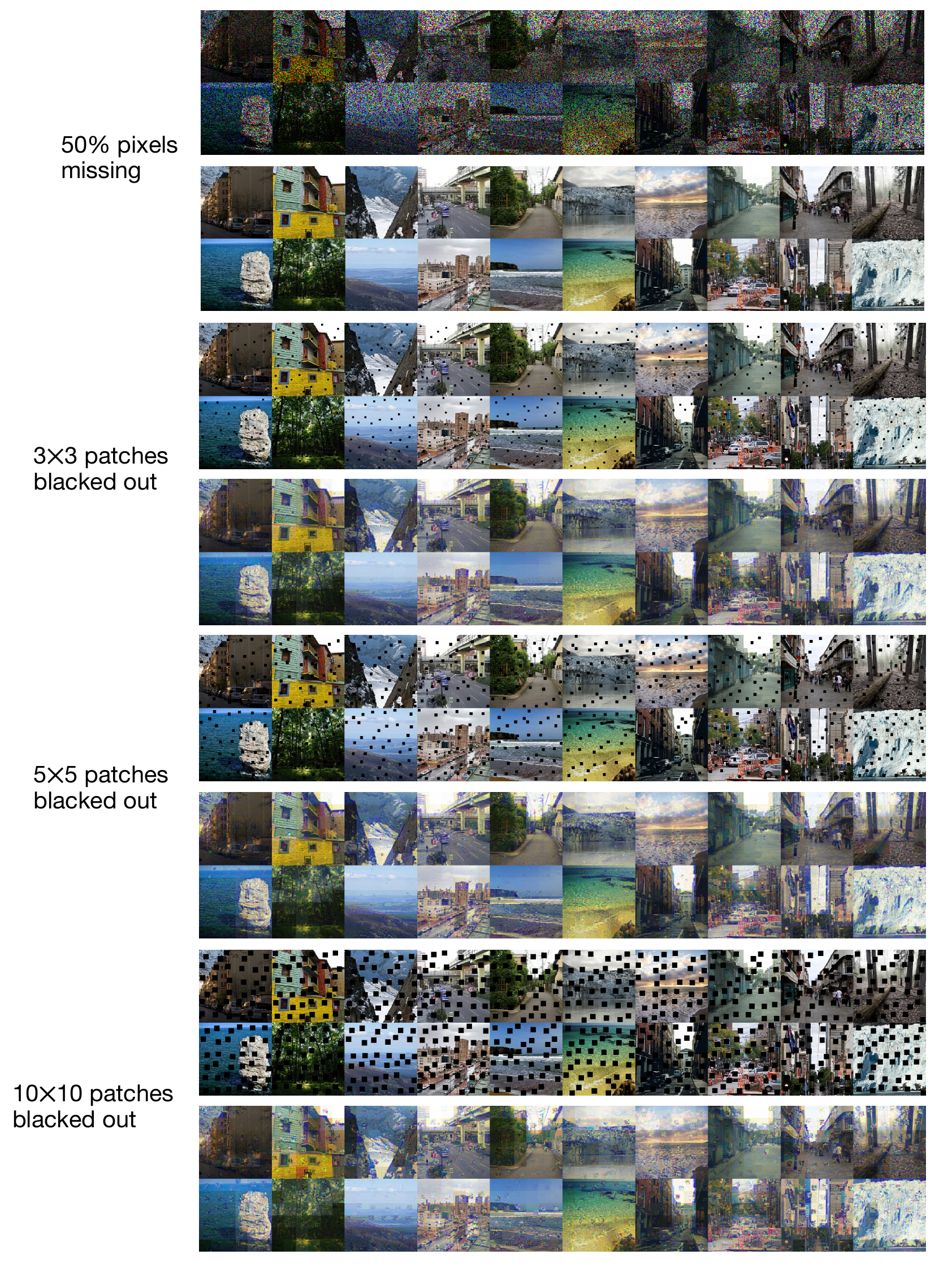}
\caption{\textbf{Image denoising and inpainting}. Given a dataset of multiple landscape images, we divided each image into patches and combined them to form a large data matrix. A dictionary is learned using our mini-batch alternating minimization algorithm. A total of 20 noisy images are shown with different noise patterns, alongside their respective reconstructed versions.}
\label{fig:block}
\end{figure}

For both the image denoising examples in Figure \ref{fig:denoise} and Figure \ref{fig:block}, we set $p_1 = p_2 = 500$ in Algorithm \ref{alg: complete mini-batch} (corresponding to a batch size of 500), and run it for a total of $T = 2000$ iterations, which took $125$ seconds for the constructed data matrix $\bY$. We randomly sample 20 images from the dataset, which are then corrupted with noise. The images before and after the denoising are shown in the first two rows of Figure \ref{fig:block}. Here we see that our learned dictionary reconstructs the original image almost exactly. 

\begin{figure}
    \centering
    \includegraphics[width=0.35\textwidth ]{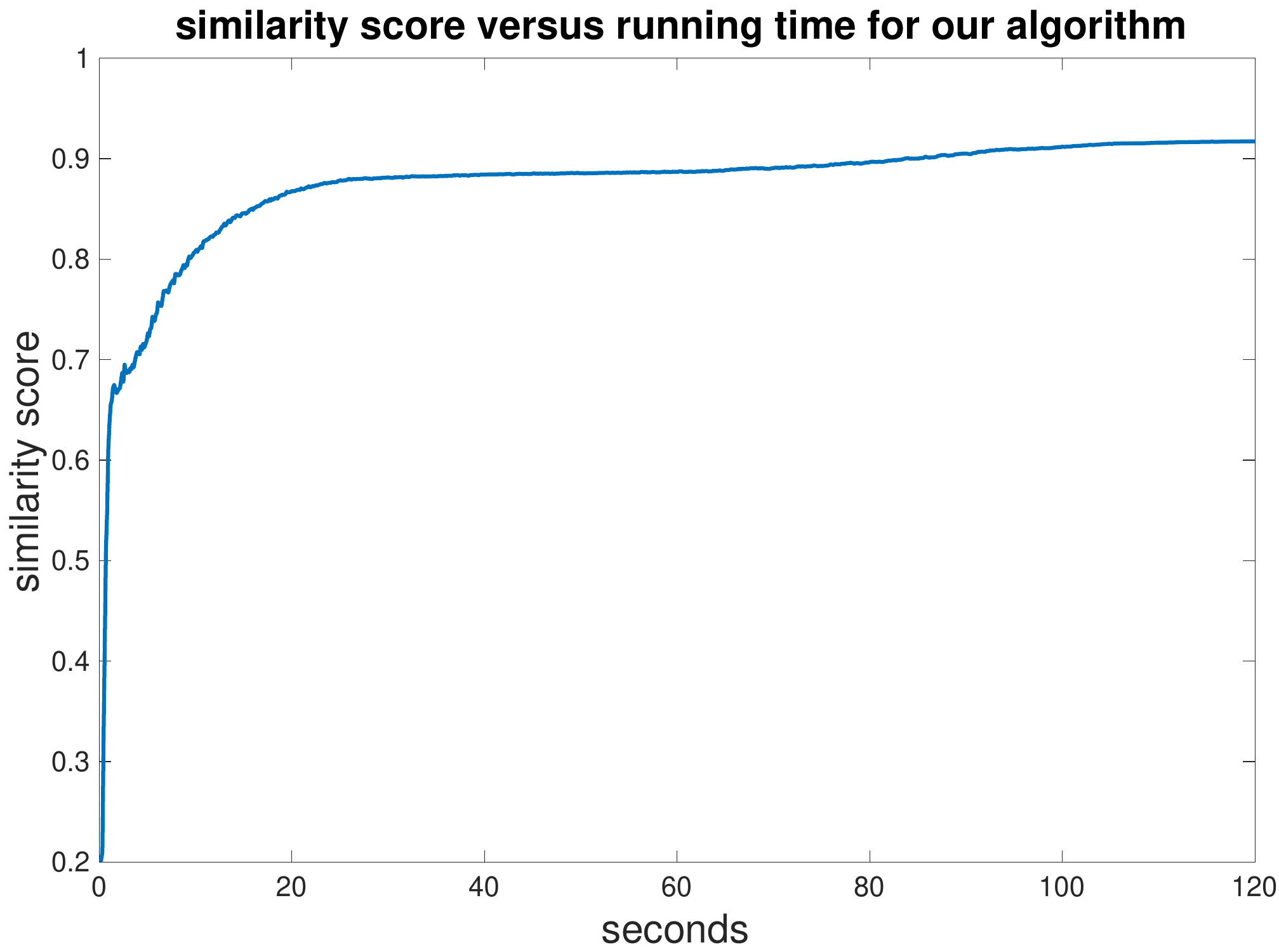}
    \includegraphics[width=0.35\textwidth]{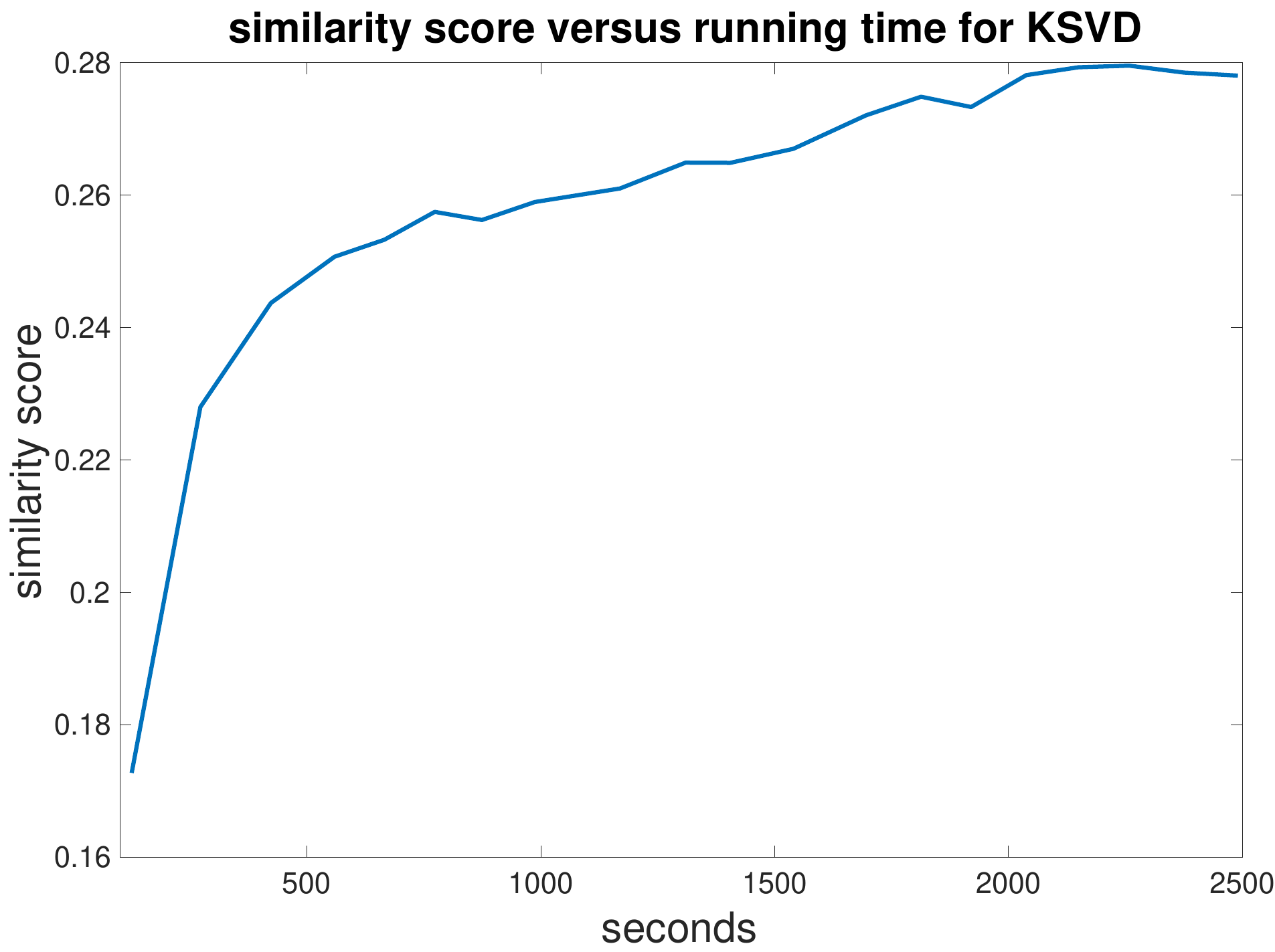}
    \caption{\textbf{Similarity score between reconstructed image and original image} We display the similarity score between the denoised image and the original one for our algorithm (left) and KSVD (right). As time progresses, the dictionary obtained from our algorithm improves, leading to increasingly better reconstructions. On the other hand, while the dictionary obtained from KSVD improves with time, the reconstruction quality exhibits a slower rate of improvement.}
    \label{fig:runtime}
\end{figure}

Additionally, we compare our learned dictionary with that of KSVD, which is implemented using a standard sparse learning library \cite{mairal2007sparse} in Matlab. 
We allow KSVD a total running time of $1100$ seconds. 
A comparison of denoised results using dictionaries learned with Algorithm \ref{alg: complete mini-batch} and KSVD are shown in Figure \ref{fig:denoise}.  Here, we also see that the dictionary learned by Algorithm \ref{alg: complete mini-batch} greatly outperforms KSVD, achieving a much better reconstruction of the original image despite being 10 times faster. Furthermore, we plot the similarity score between the reconstructed image and the original image versus the running time of both algorithms.  In particular, at each iteration of both algorithms, we use the dictionary at that iteration to reconstruct a denoised image and compare it with the original. We see in Figure \ref{fig:runtime} that our algorithm achieves a similarity score of $0.9$ in 120 seconds, while KSVD can only achieve a similarity score of 0.28 in 2500 seconds.

\paragraph{Image Inpainting}
To further gauge the efficacy of our learned dictionary, here we consider the more challenging task of \textit{inpainting}. Here instead of corrupting each image with 50\% of missing pixels, we opt to blackout entire pixel blocks.
We segment each image into patches with dimensions of $25\times 25$ to facilitate inpainting for larger areas of obscured pixels. Much like our previous approach, we execute Algorithm \ref{alg: complete mini-batch} for a total of $T=2,000$, which consumed approximately 2,100 seconds. For the reconstruction process, we maintain a sparsity level of approximately 10\%, as we did previously.

Our reconstruction results are depicted in Figure \ref{fig:block}. We randomly blackout blocks of dimensions $3\times 3$, $5\times 5$, and $10\times 10$. In all three scenarios, the total count of obscured blocks equals the total number of patches. It is evident that for smaller blacked-out blocks, our dictionary performs well in reconstructing the images. However, when dealing with larger $10\times 10$ blocks, the inpainting task becomes more challenging. In this case, we observe that the reconstructed images exhibit some imperfections within the blacked-out regions.

\section{Conclusion}
In this paper, we study the dictionary learning problem, where the goal is to represent a given set of data samples as linear combinations of a few atoms from a learned dictionary. The existing algorithms for dictionary learning often lack scalability or provable guarantees. 
This paper shows that a simple alternating minimization algorithm provably solves both orthogonal and complete dictionary learning problems. Unlike other provably convergent algorithms for dictionary learning, our proposed method does not rely on any convex relaxation of the problem, and can be easily implemented in realistic scales. Through synthetic and realistic experiments on image denoising, we showcase the superiority of our proposed algorithm compared to the most commonly-used algorithms for dictionary learning.

\clearpage
\bibliographystyle{apalike}
\bibliography{siam/references}
\clearpage
\appendix

\section{Preliminary}\label{app_pre}
To prove our main theorems, we will rely on several preliminary results from high-dimensional statistics and matrix perturbation theory, which will be essential for our arguments.
We denote the sub-Gaussian norm and $L^2$-norm of a random variable with $\|\cdot\|_{\psi_2}$ and $\|\cdot\|_{L^2}$ respectively.
\begin{theorem}[Concentration of sample covariance matrix \cite{vershynin2018high}]\label{thm: covariance estimation}
 Let $\bx$ be a sub-Gaussian random vector in $\mathbb{R}^n$ with covariance matrix $\bSigma$, such that
\begin{equation}\nonumber
    \|\langle \bx,\bz \rangle\|_{\psi_2}\le C_{ce}\|\langle \bx,\bz \rangle\|_{L^2}\quad\text{for any }\bz\in\mathbb{R}^n,
\end{equation}
for some $C_{ce}\ge 1$. Let $\bX\in\mathbb{R}^{n\times p}$ be a matrix whose columns have identical and independent distribution as $\bx$. Then, for any $u\ge 0$ and with probability at least $1-2\exp{(-u)}$, we have
\begin{equation}\nonumber
    \left\|\frac{1}{p}\bX \bX^\top- \bSigma\right\|_2 \lesssim C_{ce}^2\left(\sqrt{\frac{n+u}{p}}+\frac{n+u}{p}\right)\|\bSigma\|_2
\end{equation} 
\end{theorem}
.
\begin{theorem}[Concentration of norm \cite{vershynin2018high}]\label{thm: norm concentration}
Let $\bx \in \mathbb{R}^n$ be a random vector with independent, sub-Gaussian coordinates $\bx_i$ that satisfy $\mathbb{E} \bx_i^2=$ 1. Then, $\|\| \bx\left\|_2-\sqrt{n}\right\|_{\psi_2} \leq C K^2$,
where $K=\max _i\left\|\bx_i\right\|_{\psi_2}$ and $C$ is an absolute constant.
\end{theorem}
We next introduce a perturbation bound for the polar decomposition.
\begin{theorem}[Perturbation bound for polar decomposition \cite{li1995new}]\label{thm: perturbation bound for polar decomposition}
    Let $\bA, \tilde \bA\in \mathbb{R}^{n\times n}$ be full rank matrices with polar decompositions $\bA=\bU\bP$ and $\tbA=\tbU\tbP$. Then
    \begin{equation}\nonumber
        \|\bU-\tbU\|_F\le \frac{2}{\sigma_n(\bA)+\sigma_n(\tbA)}\|\bA-\tbA\|_F.
    \end{equation}
\end{theorem}

\section{Proof of Auxiliary Lemmas}\label{app_lemmas}
\subsection{Proof of Lemma~\ref{lemma: covariance estimation on X}}\label{proof_lem_B4}
Consider each column vector of $\bX^*$ as a random vector. Upon defining $\Sigma_{\bX^*}$ as the covariance matrix of $\bx$, we have $\Sigma_{\bX} = \theta\sigma^2\bI_n$.
We now use Theorem~\ref{thm: covariance estimation} to prove Lemma~\ref{lemma: covariance estimation on X}. For any unit-norm $\bz\in\mathbb{R}^{n}$ and any $i$th column of $\bX^*$, we have
\begin{align}\nonumber
    \left\|\langle \bX^*_{(\cdot,i)},\bz\rangle\right\|^2_{\psi_2} &= \left\|\sum_{j=1}^{n} \bX^*_{(j,i)}z_j\right\|^2_{\psi_2}
    \leq \sum_{j=1}^{n} z_j^2\left\|\bX^*_{(j,i)}\right\|^2_{\psi_2}\leq C_1\sigma^2,
\end{align}
for some constant $C_1>0$.The last inequality follows from the fact that, for a sub-Gaussian random variable, the sub-Gaussian norm is always bounded by its variance. Moreover, 
\begin{equation}\nonumber
    \left\|\langle \bX^*_{(\cdot,i)},\bz\rangle\right\|^2_{L^2}=
    \mathbb{E}\langle  \bX^*_{(\cdot,i)},\bz\rangle^2 = \bz^\top \Sigma_{\bX^*} \bz = \theta\sigma^2.
\end{equation}
We are now ready to invoke Theorem~\ref{thm: covariance estimation}. By choosing $C_{ce}=C_1/\sqrt{\theta}$, we have for some constant $C_2>0$:
\begin{equation}\label{eq_XX}
    \left\|\frac{1}{p}\bX^* {\bX^*}^\top - \theta\sigma^2\bI_n\right\|_2 \le  C_2\left(\sqrt{\frac{n+u}{p}}+\frac{n+u}{p}\right)\sigma^2,
\end{equation}  
with probability $1-2\exp{(-u)}$. Upon setting $u=n$ and assuming $p\ge  8C_2^2\tau n/\theta^2$, one can bound the right hand side by
\begin{equation}\nonumber
    C_2\left(\sqrt{\frac{n+u}{p}}+\frac{n+u}{p}\right)\sigma^2\le 2C_2\sqrt{\frac{2n}{p}}\sigma^2\le \frac{\theta\sigma^2}{\tau^{1/2}}.
\end{equation}
Combining the above inequality with~\eqref{eq_XX} leads to the desired inequalities.
$\hfill\square$
\subsection{Proof of Lemma~\ref{lemma: tilde Y to real Y}}\label{proof_lem_B3}
Let us define $p=p_1+t$. We start by noting that 
\begin{equation}\nonumber
    \frac{1}{p\theta \sigma^2}\bY \bY^\top = \bA^*\left(\frac{1}{p\theta \sigma^2}\bX^*\bX^{*\top}\right) \bA^{*\top}.
\end{equation}
Moreover, according to Theorem~\ref{thm: covariance estimation}, we have
\begin{align}\label{eq: X* spec norm}
    \left\|\frac{1}{p}\bX^*\bX^{*\top}  - \theta\sigma^2\bI_{n\times n}\right\|_2 \lesssim  \sigma^2\sqrt{\frac{n+u}{p}}
\end{align}
with probability at least $1-2\exp(-u)$. With the same probability, we have
\begin{align}\nonumber
    \left\|\frac{1}{p\theta\sigma^2}\bY\bY^\top -\bA^*\bA^{*\top} \right\|_2
    =\ &\left\|\bA^* \left(\frac{1}{p\theta \sigma^2}\bX^*\bX^{*\top}-\bI_{n\times n}\right)\bA^{*\top} \right\|_2
    \leq\ \frac{C_2}{\theta}\sqrt{\frac{n+u}{p}}
\end{align}
where in the last inequality, we used $\norm{\bA^*}_2=1$.
We define $\bDelta_1 = \frac{1}{p\theta\sigma^2}\bY\bY^\top -\bA^*\bA^{*\top} $. Using the Taylor expansion, we have 
\begin{align}
    &\left(\frac{1}{p\theta\sigma^2}\bY\bY^\top\right)^{-1}
    =\ \left(\bA^*\bA^{*\top}+\bDelta_1\right)^{-1}\nonumber\\
    &=  \left(\left(\bA^*\bA^{*\top}\right)^{1/2}\left(I_{r\times r}+\left(\bA^*\bA^{*\top}\right)^{-1/2}\bDelta_1\left(\bA^*\bA^{*\top}\right)^{-1/2}\right)\left(\bA^*\bA^{*\top}\right)^{1/2}\right)^{-1}\nonumber\\
    &=\left(\bA^*\bA^{*\top}\right)^{-1}-\left(\bA^*\bA^{*\top}\right)^{-1}\bDelta_1\left(\bA^*\bA^{*\top}\right)^{-1}+O(\bDelta_1^2)\nonumber
\end{align}
As a result, we have
\begin{equation}\nonumber
    \left\|\left(\frac{1}{p\theta\sigma^2}\bY\bY^\top\right)^{-1}-\left(\bA^*\bA^{*\top}\right)^{-1}\right\|_2
    C_3 \frac{\hkappa^4}{\theta}\sqrt{\frac{n+u}{p}}.
\end{equation}
for some constant $C_3>0$. Similarly, by Corollary 4.8 from \cite{bhatia1994matrix}, we have 
\begin{align}
    &\left\|\mathcal{L}\left(\left(\frac{1}{p\theta\sigma^2}\bY\bY^\top\right)^{-1}\right)-\mathcal{L}\left(\left(\bA^*\bA^{*\top}\right)^{-1}\right)\right\|_2\nonumber \\
    & \leq C_3\kappa\left(\left(\bA^*\bA^{*\top}\right)^{-1}\right)\left\|\left(\frac{1}{p\theta\sigma^2}\bY\bY^\top\right)^{-1}-\left(\bA^*\bA^{*\top}\right)^{-1}\right\|_2
    \le C_3\frac{\hkappa^6}{\theta}\sqrt{\frac{n+u}{p}}.\nonumber
\end{align}
We conclude the proof by setting $u= n$.$\hfill\square$
\subsection{Proof of Lemma~\ref{lemma: improvement on polar}}\label{proof_lem_B2}
Upon choosing $\tau=1$ in Lemma~\ref{lemma: covariance estimation on X}, we have $\kappa\left(\bX^*\right)<{11}/{9}$ with probability $1-2\exp\left\{-n\right\}$. Then, by setting  $\bE=\left(\bX-\bX^*\right)/\|\bX^*\|_2$ and $\hbX^*=\bX^*/\|\bX^*\|_2$, we immediately have 
\begin{equation}\label{R^s+1-R^*bound}
    \|\bD - \bD^*\|_F\le\frac{\kappa(\bX^*)^4}{2}\left\|\hbX^{*}\bE^\top-\bE \hbX^{*\top}\right\|_F+C_T\|\bE\|^2_F
\end{equation}
after invoking Lemma~\ref{q_lm}. Now define $\mathcal{T}(\cdot)$ to be the operator that replaces all the diagonal entries of a matrix with zeros. It is easy to see that $    \frac{1}{2}\left\| \hbX^{*}\bE^\top-\bE \hbX^{*\top}\right\|_F
    \le \left\|\mathcal{T} (\hbX^{*}\bE^\top)\right\|_F$.
To further investigate this bound, we introduce two matrices. Let matrix $\bM_{k}$ denote an $n \times n$ diagonal matrix of ones and a zero at location $(k, k)$. Left multiplying $\hbX^*$ by $\bM_{k}$ corresponds to replacing the $k$th row of $\hbX^*$ with zeros. Let $\tilde{\bM}_{k}$ denote an $p \times p$ diagonal matrix that has ones at entries $(i, i)$ for $i \in \supp\left(\hbX^{*}_{(k,\cdot)}\right)$ and zeros elsewhere. Right multiplying $\tbX^*$ by $\tilde{\bM}_{k}$ corresponds to replacing all the columns that are zero at $k$th row with zeros. Now, we make the following observation:
\begin{align}
    \left\|\mathcal{T} (\hbX^{*}\bE^\top)\right\|_F
    &=\sqrt{\sum_{k=1}^n \left\|(\mathcal{T} (\hbX^{*}\bE^\top))_{(\cdot,k)}\right\|_2^2}=\sqrt{\sum_{k=1}^n \left\|\left(\bM_{k}\hbX^*\tilde{\bM}_{k}\right) \bE_{(k,\cdot)}\right\|_2^2}\nonumber \\
    &\le \sqrt{\sum_{k=1}^n \left\|\bM_{k}\hbX^*\tilde{\bM}_{k}\right\|_2^2 \left\|\bE_{(k,\cdot)}\right\|_2^2}\le \max_{1\le k \le n} \left\|\bM_{k}\hbX^*\tilde{\bM}_{k}\right\|_2 \|\bE\|_F.\nonumber
\end{align}
The second equality is due to $\supp(\bE)\subseteq \supp(\hbX^*)$, and that the normalization step $\hbX^*=\bX^*/\|\bX^*\|_2$  does not change the support of $\hbX^*$. We first focus on $\max_{1\le k \le n} \left\|\bM_{k}\bX^*\tilde{\bM}_{k}\right\|_2$. Define $\bG^k\in\mathbb{R}^{(n-1)\times p}$ as the matrix $\bM_{k}\bX^*\tilde{\bM}_{k}$ after removing its $k$th row. One can see that $\|\bG^k\|_2=\|\bM_{k}\bX^*\tilde{\bM}_{k}\|_2$. Without loss of generality, we assume $k=n$, which means that we remove the last row. Recall that right multiplying $\bX^*$ by $\tilde{\bM}_{k}$ will replace each column of $\bX^*$ whose $k$th row is zero by an all-zero column vector. Therefore, $\bG^k$ has the following property:
\begin{equation}\nonumber
    \mathbbm{1}_{\bG^k_{ij}\ne 0} = B_{kj}B_{ij}\quad\text{where}\quad B_{kj}, B_{ij}\overset{i.i.d.}{\sim}\mathcal{B}(\theta) \text{ for $1\le i \le n-1$, $1\le j\le p$ }.
\end{equation}
In short, $\bG^k$ is a matrix that satisfies Assumption~\ref{assump:sparse and normalization} with parameter $\theta^2$. Therefore, we can invoke Lemma~\ref{lemma: covariance estimation on X} to bound $\|\bG^k\|_2$ for each $k$. Given $p\gtrsim \tau (n+u)/\theta^2$, we have that $\|\bG^k\|_2\le \left(1+\tau^{-1/4}\right)\sqrt{p}\theta\sigma$ for some specific $k$,
with probability $1-2\exp(-u)$. To bound maximal $\|\bG^k\|_2$ for $1\le k \le n$, we take the union bound and obtain
\begin{equation}\label{eq: tail bound DXD}
    \max_{1\le k \le n} \left\|\bM_{k}\bX^*\tilde{\bM}_{k}\right\|_2 = \max_{1\le k \le n}\|\bG^k\|_2 \le\left(1+\tau^{-1/4}\right)\sqrt{p}\theta\sigma,
\end{equation}
with probability $1-2\exp(\log n-u)$ as long as $p\gtrsim \tau (n+u)/\theta^2$. Combining this result with~\eqref{R^s+1-R^*bound}, we have
\begin{align}
    &\|\bD - \bD^*\|_F
    \le\frac{\kappa(\bX^*)^4}{2}\left\|\hbX^{*}\bE^\top-\bE \hbX^{*\top}\right\|_F+C_T\|\bE\|^2_F\nonumber\\
    &\le \frac{\kappa(\bX^*)^4\max_{1\le k \le n} \left\|\bM_{k}\bX^*\tilde{\bM}_{k}\right\|_2}{\|\bX^*\|_2} \|\bE\|_F+C_T\|\bE\|^2_F\nonumber\\
    &\!\overset{(a)}{\le}\! \frac{\left(\frac{1+\tau^{-1/4}}{1-\tau^{-1/4}}\right)^4\left(1+\tau^{-1/4}\right)\sqrt{p}\theta\sigma}{\left(1-\tau^{-1/4}\right)\sqrt{p\theta}\sigma} \|\bE\|_F\!+\!C_T\|\bE\|^2_F
    \!\le\! \left(\left(\frac{1+\tau^{-1/4}}{1-\tau^{-1/4}}\right)^5\sqrt{\theta}\!+\!C_T\|\bE\|_F\right)\|\bE\|_F,\!\nonumber
\end{align}
with probability $1-2\exp(\log n-u)$. Inequality (a) is due to Lemma~\ref{lemma: covariance estimation on X} and~\eqref{eq: tail bound DXD}. Recall that $\theta\leq 1/2$. Therefore, for sufficiently large constant $\tau$, we have $\left(\frac{1+\tau^{-1/4}}{1-\tau^{-1/4}}\right)^5\sqrt{\theta} \leq 0.8$. Moreover, $C_T\|\bE\|_F\leq 0.1$, provided that $\norm{\bE}_F\leq 1/(10C_T)$. Therefore, we have $\|\bD - \bD^*\|_F\leq 0.9\norm{\bE}_F$. The proof is complete upon choosing $u=n$. $\hfill\square$

\subsection{Proof of Lemma~\ref{lemma: exact support recovery}}\label{proof_lem_B1} The $(i,j)$-th entry $\left( \bD^\top\bY\right)_{ij}$ of $\bD^\top\bY$ can be written as
\begin{align}\label{eq: decompose of an entry}
    &\left( \bD^\top\bY\right)_{ij} = \langle \bD_{(\cdot,i)},\bY_{(\cdot,j)} \rangle= \underbrace{\langle \bD_{(\cdot,i)},\bD^*_{(\cdot,i)}\rangle \bX^*_{ij}}_{:=\cA_{ij}} +\underbrace{\sum_{k\ne i } \langle \bD_{(\cdot,i)},\bD^*_{(\cdot,k)}\rangle \bX^*_{kj}}_{:=\cB_{ij}}.
\end{align}
The first term $\cA_{ij}$ can be decomposed as 
\begin{align}
  \cA_{ij}
    = \langle \bD_{(\cdot,i)},\bD_{(\cdot,i)}\rangle \bX^*_{ij}+\langle \bD_{(\cdot,i)},\bD^*_{(\cdot,i)}-\bD_{(\cdot,i)}\rangle \bX^*_{ij}
    = \bX^*_{ij}+\langle \bD_{(\cdot,i)},\bD^*_{(\cdot,i)}-\bD_{(\cdot,i)}\rangle \bX^*_{ij}.\nonumber
\end{align}
As a result, when $\bX^*_{ij}=0$, we have $\cA_{ij}=0$. Moreover, when $\bX^*_{ij}\ne 0$, we have $ |\cA_{ij}|\ge (1-\|\bD^*-\bD\|_{1,2})\Gamma\ge \frac{3\Gamma}{4}$, where the last inequality holds for every $n\geq \frac{16}{\sigma^2\theta}$. The choice of the constant $3/4$ is to streamline the proof and can be replaced by any constant in $(\frac{1}{2},1)$. To prove the lemma it suffices to show that, for all $\bD$ such that $\|\bD-\bD^*\|_{1,2}\leq {C_1}/n$ for some $C_1$ to be defined later, we have $|\cB_{ij}|\leq \Gamma/4$ with high probability. Note that this will lead to $\left|\left( \bD^\top\bY\right)_{ij}\right|\ge  |\cA_{ij}|- |\cB_{ij}|\ge \Gamma/2$, thereby proving the statement.
We have
\begin{align}
    \left|\cB_{ij} \right |
    & =  \left|\left\langle \bD_{(\cdot,i)},\sum_{k\ne i }\bD^*_{(\cdot,k)}\bX^*_{kj} \right\rangle \nonumber  \right |\\
    & =  \left|\left\langle \bD_{(\cdot,i)}-\bD^*_{(\cdot,i)}, \sum_{k\ne i }\bD^*_{(\cdot,k)}\bX^*_{kj} \right\rangle \nonumber  \right |\\
    &\le \left \|\bD_{(\cdot,i)}-\bD^*_{(\cdot,i)}\right\|_2\left\|\sum_{k\ne i }\bD^*_{(\cdot,k)}\bX^*_{kj}\right\|_2\\
    &\le \left \|\bD_{(\cdot,i)}-\bD^*_{(\cdot,i)}\right\|_2\left\|\sum_{k =1 }^{n}\bD^*_{(\cdot,k)}\bX^*_{kj}\right\|_2\\
    &= \left \|\bD_{(\cdot,i)}-\bD^*_{(\cdot,i)}\right\|_2\left\|\bY_{(\cdot,j)}\right\|_2\nonumber\\
    &= \left \|\bD_{(\cdot,i)}-\bD^*_{(\cdot,i)}\right\|_2\left\|\bX_{(\cdot,j)}\right\|_2\nonumber.
\end{align}
Following the argument of Equation~\ref{eq: concentration for each column of X}, we have $\left\|\bX_{(\cdot,j)}\right\|_2\le 2\sigma\sqrt{\theta n}$ with probability at least $1-2\exp\left(-Cn\right)$. With the same probability, we have for any $\bD$ that satisfies $\|\bD-\bD^*\|_{1,2}\le \frac{\Gamma}{8\sigma\sqrt{\theta n}}$:
\begin{align*}
    \left|\cB_{ij} \right |\le \left \|\bD_{(\cdot,i)}-\bD^*_{(\cdot,i)}\right\|_2\left\|\bX_{(\cdot,j)}\right\|_2 \le \Gamma/4.
\end{align*}
Upon taking the union bound over all $j$th column of $\bX$, we have $\left|\cB_{ij} \right | < \Gamma/4$ for all $(i,j)$ with probability $1-2\exp\left(\log p-Cn\right)$. This completes the proof of the second statement.  $\hfill\square$

\begin{algorithm}[H]
	\caption{Rank-one update for the preconditioner}
	\label{alg: precon update}
	\begin{algorithmic}[1]
        \STATE{{\bf Input:} $\bL$, $\bA$, $\by$.}
        \STATE{Set $\bv=\frac{\bA\by}{\sqrt{1+\by^\top\bA\by}}$, $\bA' = \bA-\bv\bv^\top$, $\bL = \frac{\bL}{\sqrt{(p_1+t)\theta\sigma^2}}$, $\bw = \bv$, and $b = 1$.}
        \FOR{$j = 1,\dots, n$}
        \STATE{Set $\bL'_{jj} = \sqrt{\bL^2_{jj}-\frac{1}{b}\bw_j}$ and $\gamma = \bL^2_{jj} b - \bw_j^2$.}
        \FOR{$k = j+1,\dots, n$}
        \STATE{Set $\bw_k = \bw_k-\frac{\bw_j}{\bL_{jj}}\bL_{kj}$ and $\bL'_{kj} = \frac{\bL'_{jj}}{\bL_{jj}}\bL_{kj}-\frac{\bL'_{jj}\bw_j}{\gamma}\bw_k$.}
        \ENDFOR
        \STATE{Set $b = b-\frac{\bw_j^2}{\bL^2_{jj}}$}
        \ENDFOR
        \STATE{Set $\bL' = \sqrt{(p_1+t+1)\theta\sigma^2}\bL'$.}
        \RETURN{$\bL'$ and $\bA'$.}
	\end{algorithmic}
\end{algorithm}

\section{Rank-one Updates for the Preconditioner}\label{app_cholesky}
We define $$\bA = \left(\bY\bY^\top\right)^{-1},\quad \bL = \bP^{(t-1)},\quad \bA' = \left(\bY\bY^\top+\by\by^\top\right)^{-1},\quad \bL' = \bP^{(t)}.$$
The above rank-one update algorithm for Cholesky decomposition (Algorithm~\ref{alg: precon update}) is already implemented in MATLAB function \texttt{cholupdate}.

\end{document}

%% file: math_commands.tex
\usepackage{amsmath,amsfonts,bm}

\def\1{\bm{1}}

\DeclareMathAlphabet{\mathsfit}{\encodingdefault}{\sfdefault}{m}{sl}
\SetMathAlphabet{\mathsfit}{bold}{\encodingdefault}{\sfdefault}{bx}{n}

